\documentclass[nohyperref]{article}

\usepackage{microtype}
\usepackage{graphicx}
\usepackage{subfigure}
\usepackage{booktabs} 
\usepackage{multirow}
\usepackage{tabularx}
\usepackage{enumerate}
\usepackage{hyperref}

\usepackage[noend]{algpseudocode}

\usepackage[accepted]{icml2023}

\usepackage{amsmath}
\usepackage{amssymb}
\usepackage{mathtools}
\usepackage{amsthm}

\usepackage[capitalize,noabbrev]{cleveref}

\theoremstyle{plain}
\newtheorem{theorem}{Theorem}[section]
\newtheorem{proposition}[theorem]{Proposition}

\theoremstyle{definition}

\theoremstyle{remark}

\usepackage[textsize=tiny]{todonotes}

\icmltitlerunning{RankMe}

\usepackage{amsmath}
\usepackage{bm}
\usepackage[utf8]{inputenc} 
\usepackage[T1]{fontenc}    
\usepackage{hyperref}       
\usepackage{url}            
\usepackage{nicefrac}       
\usepackage{microtype}      
\usepackage{xcolor}         

\usepackage{booktabs}
\usepackage{subfigure}

\def\eqref#1{equation~\ref{#1}}

\def\1{\bm{1}}

\def\vb{{\bm{b}}}

\def\vz{{\bm{z}}}

\def\mP{{\bm{P}}}

\def\mW{{\bm{W}}}

\def\mY{{\bm{Y}}}
\def\mZ{{\bm{Z}}}

\DeclareMathAlphabet{\mathsfit}{\encodingdefault}{\sfdefault}{m}{sl}
\SetMathAlphabet{\mathsfit}{bold}{\encodingdefault}{\sfdefault}{bx}{n}

\def\sP{{\mathbb{P}}}

\DeclareMathOperator{\rank}{rank}

\usepackage{tikz}

\usepackage{hyperref}
\usepackage{url}
\usepackage{multirow}
\usepackage{enumitem}

\usepackage{amsthm}
\usepackage{amssymb}
\usepackage{mathtools}

\usepackage{varwidth}
\usepackage{enumitem}
\usepackage{tikz}
\usetikzlibrary{positioning}
\usetikzlibrary{shapes}
\usetikzlibrary{fit}
\usetikzlibrary{calc}

\usepackage{listings}
\usepackage{xcolor}
\usepackage{courier}

\definecolor{codegreen}{rgb}{0,0.6,0}
\definecolor{codegray}{rgb}{0.5,0.5,0.5}
\definecolor{codeblack}{rgb}{0.,0.,0.}
\definecolor{codepurple}{rgb}{0.58,0,0.82}
\definecolor{backcolour}{rgb}{0.95,0.95,0.92}

\lstdefinestyle{mystyle}{
    backgroundcolor=\color{backcolour},   
    commentstyle=\color{codegreen},
    keywordstyle=\color{codeblack},
    numberstyle=\tiny\color{codegray},
    stringstyle=\color{codepurple},
    basicstyle=\ttfamily\footnotesize,
    breakatwhitespace=false,         
    breaklines=true,                 
    captionpos=b,                    
    keepspaces=true,                 
    numbers=left,                    
    numbersep=5pt,                  
    showspaces=false,                
    showstringspaces=false,
    showtabs=false,                  
    tabsize=2,
    aboveskip=0pt,
    belowskip=-3pt
}

\begin{document}

\newcommand{\name}{RankMe}
\newcommand{\QG}[1]{{\textcolor{orange}{[\textbf{QG:} #1]}}}
\newcommand{\RB}[1]{{\textcolor{blue}{[\textbf{RB:} #1]}}}
\setlist[enumerate]{noitemsep, topsep=0pt,leftmargin=*}

\twocolumn[
\icmltitle{RankMe: Assessing the Downstream Performance of Pretrained Self-Supervised Representations by Their Rank}

\begin{icmlauthorlist}
\icmlauthor{Quentin Garrido}{meta,uge}
\icmlauthor{Randall Balestriero}{meta}
\icmlauthor{Laurent Najman}{uge}
\icmlauthor{Yann LeCun}{meta,nyu1,nyu2}
\end{icmlauthorlist}

\icmlaffiliation{meta}{Meta AI - FAIR}
\icmlaffiliation{uge}{Univ Gustave Eiffel, CNRS, LIGM, F-77454 Marne-la-Vallée, France}
\icmlaffiliation{nyu1}{Courant Institute, New York University}
\icmlaffiliation{nyu2}{Center for Data Science, New York University}

\icmlcorrespondingauthor{Quentin Garrido}{garridoq@meta.com}

\icmlkeywords{Machine Learning, self-supervised learning, evaluation, rank, singular values, representation learning, ICML}
\vskip 0.3in
]

\printAffiliationsAndNotice{} 

\begin{abstract}
    Joint-Embedding Self Supervised Learning (JE-SSL) has seen a rapid development, with the emergence of many method variations but only few principled guidelines that would help practitioners to successfully deploy them. The main reason for that pitfall comes from JE-SSL's core principle of not employing any input reconstruction therefore lacking visual cues of unsuccessful training. Adding non informative loss values to that, it becomes difficult to deploy SSL on a new dataset for which no labels can help to judge the quality of the learned representation. In this study, we develop a simple unsupervised criterion that is indicative of the quality of the learned JE-SSL representations: their effective rank. Albeit simple and computationally friendly, this method ---coined {\em RankMe}--- allows one to assess the performance of JE-SSL representations, even on different downstream datasets, without requiring any labels. A further benefit of RankMe is that it does not have any training or hyper-parameters to tune. Through thorough empirical experiments involving hundreds of training episodes, we demonstrate how RankMe can be used for hyperparameter selection with nearly no reduction in final performance compared to the current selection method that involve a dataset's labels. We hope that RankMe will facilitate the deployment of JE-SSL towards domains that do not have the opportunity to rely on labels for representations' quality assessment.
\end{abstract}

\section{Introduction}

Self-supervised learning (SSL) has shown great progress to learn informative data representations in recent years~\citep{chen2020simple,he2020moco, chen2020mocov2, grill2020byol, lee2021cbyol, caron2020swav, zbontar2021barlow, bardes2021vicreg, tomasev2022relicv2, caron2021dino, chen2021mocov3, li2022esvit, zhou2022ibot, zhou2022mugs,haochen2021provable,he2022masked}, catching up to supervised baselines and even surpassing them in few-shot learning, i.e., when evaluating the SSL model from only a few labeled examples. Although various SSL families of losses have emerged, most are variants of the joint-embedding (JE) framework with a siamese network architecture~\citep{bromley1994siamese}, denoted as JE-SSL for short. The only technicality we ought to introduce to make our study precise is the fact that JE-SSL has introduced some different notations to denote an input's representation. In short, JE-SSL often composes a {\em backbone} or {\em encoder} network e.g., a ResNet50 and a {\em projector} network e.g., a multilayer perceptron. This projector is only employed during training, and we refer to its outputs as {\em embeddings}, while the actual inputs' {\em representation} employed for downstream tasks are obtained at the encoder's output.

Although downstream tasks performance of JE-SSL representations might seem impressive, one pondering fact should be noted: {\em all existing methods, hyperparameters, models --- and thus performances --- are obtained by manual search involving the labels of the considered datasets}. In words, JE-SSL is tuned by monitoring the supervised performance of the model at hand.
Therefore, successfully deploying a SSL model on a new dataset relies on the {\em strong assumption} of having labels on that dataset to tune the SSL method e.g. through a linear classifier feeding on the JE-SSL representations~\citep{misra2020pirl}. This quality assessment strategy was also extended to the use of nonlinear classifiers, e.g., a $k$-nn classifier~\citep{wu2018discrimination,zhuang2019local}. 
Hence, although labels are not directly employed to compute the weight updates, they are used as a proxy. This limitation prevents the deployment of JE-SSL in challenging domains where the number of available labelled examples is limited. Adding to the challenge, one milestone of JE-SSL is to move away from reconstruction based learning; hence without labels and without visual cues, tuning JE-SSL methods on unlabeled datasets remains challenging. This led to the application of feature inversion methods e.g. Deep Image Prior~\citep{ulyanov2018deep} or conditional diffusion models~\citep{bordes2021high} to be deployed onto learned JE-SSL representation to try to visualize the learned features. Those alternative visualization solutions however suffer from their own limitations e.g. bias of the used method, or computational cost. More importantly, those feature inversion strategies have been designed for natural images i.e. it is not clear how such methods would perform on different data modalities.

\begin{figure*}[t!]
    \centering
    \includegraphics[width=1\textwidth]{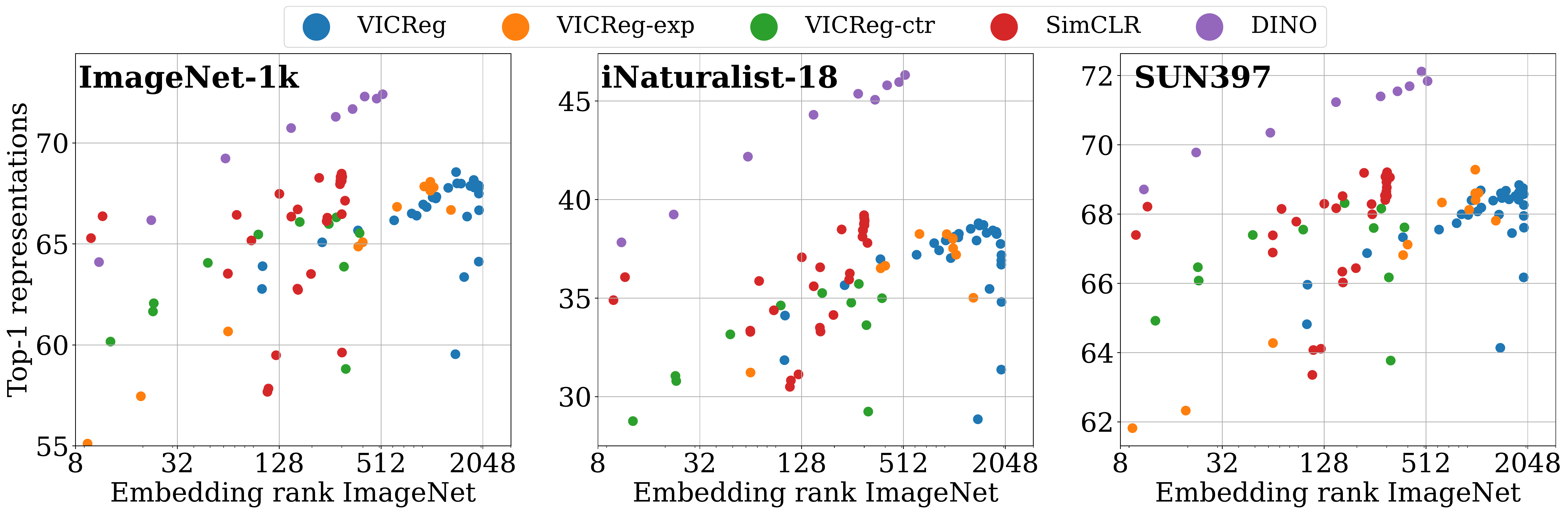}

  \begin{tabular}{llcccccccccc}
    \toprule
     \multirow{2}{*}{Dataset} & \multirow{2}{*}{Method} & \multirow{2}{*}{Labels} & \multicolumn{4}{c}{VICReg} &  \multicolumn{3}{c}{SimCLR} &  \multicolumn{2}{c}{DINO} \\ 
 \cmidrule(lr){4-7} \cmidrule(lr){8-10} \cmidrule(lr){11-12} & &  & cov. & inv. & LR & WD & temp. & LR. & WD. & t-temp. & s-temp.\\ 
 \midrule 
\multirow{3}{*}{ImageNet} & \textcolor{gray}{ImageNet Oracle} & \textcolor{gray}{\checkmark} & \textcolor{gray}{68.2} & \textcolor{gray}{68.2} & \textcolor{gray}{68.6} & \textcolor{gray}{68.0} & \textcolor{gray}{68.5} & \textcolor{gray}{68.5} & \textcolor{gray}{68.3} & \textcolor{gray}{72.3} & \textcolor{gray}{72.4}\\ 
 & $\alpha$-ReQ & X & \textbf{67.9} & 67.5 & 59.5 & \textbf{67.8} & 63.5 & \textbf{68.1} & 32.3 & 71.7 & 66.2\\ 
 & \name{} & X & 67.8 &\textbf{ 67.9} & \textbf{68.2} & \textbf{67.8} & \textbf{67.1} & 68.0 & \textbf{68.3} & \textbf{72.2} & \textbf{72.4}\\ 
  \midrule 
\multirow{3}{*}{OOD} & \textcolor{gray}{ImageNet Oracle} & \textcolor{gray}{\checkmark} &  \textcolor{gray}{68.7} & \textcolor{gray}{68.7} & \textcolor{gray}{68.9} & \textcolor{gray}{68.8} & \textcolor{gray}{68.7} & \textcolor{gray}{68.7} & \textcolor{gray}{68.8} & \textcolor{gray}{71.9} & \textcolor{gray}{72.5}\\ 
 & $\alpha$-ReQ & X & \textbf{68.1} & 67.8 & 63.8 & \textbf{68.4} & 65.1 & 68.2 & 68.6 & \textbf{71.8} & 68.5\\ 
 & \name{} & X & 67.7 & \textbf{68.3} & \textbf{68.7} & \textbf{68.4} &\textbf{ 67.6} & \textbf{68.4} & \textbf{68.8} & \textbf{71.8} & \textbf{72.5}\\  
    \bottomrule
  \end{tabular}

    \caption{
    \textbf{Top:} Performance of JE-SSL representations (encoder output) in {\bf y-axis} against the embeddings (projector output) \name{} values in {\bf x-axis} on ImageNet-1k. Except for some degenerate solutions at full-rank, \name{} values correlate well with in-distribution ({\bf left column}) and out-of-distribution ({\bf right columns}) classification performance. \textbf{Bottom:} Hyperparameter selection using the common supervised linear probe strategy, $\alpha$-ReQ the proposed unsupervised \name{} strategy. Values in bold represent the best performance between \name{} and $\alpha$-ReQ.  OOD indicates the average performance over all the considered datasets other than ImageNet. Without any label, optimization or parameters, \name{} is able to recover most of the performance obtained by using ImageNet validation set, highlighting its strength as a hyperparameter selection tool. \name{} also outperforms $\alpha$-ReQ on average and does not suffer from as big performance drops in worst cases.}
    \label{fig:figure_1}
    \vspace{-0.4cm}
\end{figure*}

In this study we propose \name{} to assess a model's performance without having access to any labels; a simple method that does not require any training or tuning. \name{} accurately predicts a model's performance both on In-Distribution (ID), i.e., same data distribution as used during the JE-SSL training, and on Out-Of-Distribution (OOD), i.e., different data distribution onto which the learned model is deployed onto. We highlight this  crucial property at the top of \cref{fig:figure_1}. The strength of \name{} lies in the fact that it is solely based on the singular values distribution of the learned embeddings which is not only simple to obtain but also easy to interpret. In fact, \name{}'s motivation hinges on Cover's theorem~\citep{cover1965geometrical} that states how increasing the rank of a linear classifier's input increases its training performance, and three simple hypotheses that thoroughly validate empirically at the end of our study. Since \name{} provides a step towards (unlabeled) JE-SSL by allowing practitioners to cross-validate hyperparameters and select models without resorting to labels or feature inversion methods, we hope that it will allow JE-SSL to move away from using labels as part of their design search strategy. We summarize our contributions below:
\begin{enumerate}
    \item We introduce RankMe (\cref{eq:rank_smooth}) and motivate its construction from first principles (\cref{sec:theory}) e.g. Cover's theorem
    \item We demonstrate that \name{}'s ability to inform about JE-SSL downstream performances is consistent across methods, e.g. VICReg, SimCLR, DINO, and their variants, and across architectures, e.g. using a projector network and/or a nonlinear evaluation method (see \cref{fig:perfs_repr,sec:nonlinear})
    \item We demonstrate that \name{} enables hyperparameter cross-validation for JE-SSL methods; \name{} is able to retrieve and sometimes surpass most of the performance previously found by manual --and label-guided-- search while not employing any labels, on both in domain and out of domain datasets (\cref{fig:figure_1,tab:hp-tuning,tab:inat-pretraining})
\end{enumerate}

We provide a hyperparameter free numerically stable implementation of \name{} in \cref{sec:method-presentation} and pseudo-code for cross-validation in \cref{fig:hp-selection}. Through extensive experiments involving 11 datasets and 110 models over 5 methods, we demonstrate that in the linear and nonlinear probing regime, \name{} is able to tell apart successful and sub-optimal JE-SSL training, even on different downstream tasks without having access to labels or downstream task data samples.

\vspace{-0.2cm}
\section{Background}
\vspace{-0.2cm}

{\bf Joint embedding self-supervised learning (JE-SSL).} In JE-SSL, two main families of method can be distinguished: contrastive and non-contrastive. Contrastive methods~\citep{chen2020simple,he2020moco,chen2020mocov2,chen2021mocov3,yeh2021decoupled} mostly rely on the InfoNCE criterion~\citep{oord2018infonce} except for~\cite{haochen2021provable} which uses squared similarities between the embedding. A clustering variant of contrastive learning has also emerged~\citep{caron2018clustering,caron2020swav,caron2021dino} and can be thought of as contrastive methods, but between cluster centroids instead of samples. Non-contrastive methods~\citep{grill2020byol,chen2020simsiam,caron2021dino,bardes2021vicreg,zbontar2021barlow,ermolov2021whitening,li2022neural} aim at bringing together embeddings of positive samples, similar to contrastive learning. However, a key difference with contrastive learning lies in how those methods prevent a representational collapse. In the former, the criterion explicitly pushes away negative samples, i.e., all samples that are not positive, from each other. In the latter, the criterion does not prevent collapse by distinguishing positive and negative samples, but instead considers the embeddings as a whole and encourages information content maximization e.g., by regularizing the empirical covariance matrix of the embeddings. Such a categorization is not needed for our development, and we thus refer to any of the above method as JE-SSL. 

{\bf Known Observations About Representations' Spectrum in JE-SSL.} The phenomenon of learning rank-deficient or dimensional collapsed, embeddings in JE-SSL has recently been studied from both a theoretical and empirical point of view. The empirical emergence of dimensional collapse was studied in~\cite{hua2021feature} where they proposed the use of a whitening batch normalization layer to help alleviate it. In~\cite{jing2022understanding}, a focus on contrastive approaches in a linear setting enabled a better understanding of dimensional collapse and the role of augmentations in its emergence. Performance in a low label regime of a partially collapsed encoder can also be improved by forcing the whitening of its output, as shown in~\cite{he2022exploring}. Furthermore, it was shown in~\cite{balestriero2022spectral} how dimensional collapse is a phenomenon that should not necessarily happen in theory and how its emergence is mostly due to practical concerns. Interestingly, we will see through the lens of \name{} that dimensional collapse is tightly linked with the quality of the representation. In supervised learning, the collapse of the embeddings was also studied and found to be detrimental to performances~\cite{ganea2019breaking}.

As such, existing studies have started to prescribe informally the choice of representations that have a lesser collapse; yet no formal study on the ability of this recipe to actually identify successfully trained models, nor how to quantify the amount of collapse to improve representations as been proposed; this is the goal of our study.

\vspace{-0.2cm}
\section{\name{} Consistently Predicts Downstream performances From Representations}
\vspace{-0.2cm}

The goal of this section is to introduce and motivate \name{} while providing a numerically stable implementation. We defer a theoretical justification to \cref{sec:theory}.
To ease notations, we refer to the (train) dataset used to obtain the JE-SSL model as {\em source dataset}, and the test set on the same dataset or a different OOD dataset as {\em target dataset}.

\subsection{\name{}: A Simple Method and Its Implementation\label{sec:method-presentation}}
The most crucial step of \name{} is the estimation of the embeddings' rank. A trivial solution could be to check at the number of nonzero singular values. Denoting by $\sigma_k$ the $k^{\rm th}$ singular value of the $(N \times K)$ embedding matrix $\mZ$, this would lead to $\rank(\mZ)=\sum_{k=1}^{\min(N,K)}1_{\{\sigma_k > 0\}}$. However, such a definition is too rigid for practical scenarios. For example, round-off error alone could have a dramatic impact on the rank estimate.
Instead, alternative and robust rank definitions have emerged \citep{press2007numerical} such as
    \mbox{$\rank(\mZ)=\sum_{k=1}^{\min(N,K)}1_{\{\sigma_k > \max_{i}\sigma_i \times  \max(M, N) \times \epsilon\}},\label{eq:rank}$}
where $\epsilon$ is a small constant dependent on the data type, typically $~10^{-7}$ for \texttt{float32}.
An alternative measure of rank comes from a probabilistic viewpoint where the singular values are normalized to sum to 1 and the Shannon Entropy~\citep{shannon1948mathematical} is used, which corresponds to our definition of \name{} from \cref{eq:rank_smooth}. 
We thus introduce \name{} formally as the following smooth rank measure, originally introduced in~\cite{roy2007effective},
\begin{align}
    &\text{\name{}}(\mZ) = \exp\left(-\sum_{k=1}^{\min(N,K)} p_k \log p_k\right),\label{eq:rank_smooth}
    \\
    & \text{with}\quad p_k = \frac{\sigma_k(\mZ)}{\|\sigma(\mZ)\|_1}+\epsilon,
\end{align}
where $\mZ$ is the source dataset's embeddings. 
As opposed to the classical rank, the chosen~\cref{eq:rank_smooth} does not rely on specifying the exact threshold at which the singular value is treated as nonzero. Throughout our study, we employ \cref{eq:rank_smooth}, and provide the matching analysis with the classical rank in the appendix. Another benefit of \name{}'s \cref{eq:rank_smooth} is in its quantification of the whitening of the embeddings in addition to their rank, which is known to simplify optimization of (non)linear probes put on top of them~\citep{santurkar2018does}.
Lastly, although \cref{eq:rank_smooth} is defined with the full embedding matrix $\mZ$, we observe that not all of the samples need to be used to have an accurate estimate of \name{}. In practice, we use $25 600$ samples as ablation studies provided in \cref{sec:rank-convergence} and \cref{fig:rank-convergence} indicate that this provides a highly accurate estimate.
\name{} should however only be used to compare different runs of a given method, since the embeddings' rank is not the only factor that affects performance.

\textbf{Relation of \name{} To Existing Solutions.}
Performance evaluation without labels can also be done using a pretext-task, such as rotation prediction. This technique helped in selecting data augmentation policies in ~\cite{reed2021selfaugment}. One limitation lies in the need to select and train the classifier of the pretext-task, and on the strong assumption that rotation were not part of the transformations one aimed to be invariant to.
Since (supervised) linear evaluation is the most widely used evaluation method, we will focus on showing how \name{} compares with it. In~\cite{li2022understanding}, it is shown that the eigenspectrum of representations can be used to assess performance when used in conjunction with the loss value. This requires training an additional classifier to predict the performance and as such is not usable as is in a completely unsupervised fashion.
Most related to us is~\cite{ghosh2022investigating} where representations are evaluated by their eigenspectrum decay, giving a baseline for unsupervised hyperparameter selection.
$\alpha$-ReQ relies on strong assumptions, and if they hold, then \name{} and $\alpha$-ReQ can match, but we show that we outperform it on average. In fact the assumptions made by $\alpha$-ReQ are known to not hold in light of collapse~\citep{he2022exploring}. We investigate $\alpha$-ReQ's behavior in detail in \cref{sec:areg}.

\subsection{\name{} Predicts Linear Probing performance Even on Unseen Datasets}
\label{sec:ood_rank_perf}
\label{sec:ood_rank}

In order to empirically validate \name{}, we compare it to linear evaluation, which is the default evaluation method of JE-SSL methods. Finetuning has gained in popularity with Masked Image Modeling methods~\cite{he2021mae}, but this can have a significant impact on the properties of the embeddings and alters what was learned during the pretraining. As such, we do not focus on this evaluation.

{\bf Experimental Methods and Datasets Considered.}~
In order to provide a meaningful assessment of the embeddings rank's impact on performance, we focus on 5 JE-SSL methods. We use SimCLR as a representative contrastive method, VICReg as a representative covariance based method, and VICReg-exp and VICReg-ctr which were introduced in~\cite{garrido2022duality}. We also include DINO~\cite{caron2021dino} as a clustering approach. Applying \name{} to DINO is not as straightforward due to the clustering layer in the projector, so embeddings have to be taken right before the last projector layer. Confer \cref{sec:dino} for more details. To make our work self-contained, we present the methods in \cref{sec:background}. We chose to use VICReg-exp and VICReg-ctr as they provide small modifications to VICReg and SimCLR while producing embeddings with different rank properties. For each method we vary parameters that directly influence the rank of the embeddings, whether it is the temperature used in softmax based methods, which directly impacts the hardness of the softmax, or the loss weights to give more or less importance to the regularizing aspect of loss functions. We also vary optimization parameters such as the learning rate and weight decay to provide a more complete analysis. We provide the hyperparameters used for all experiments in \cref{sec:all-tables}. 
All approaches were trained in the same experimental setting with a ResNet-50~\citep{he2016resnet} backbone with a MLP projector having intermediate layers of size $8192,8192,2048$, which avoids any architectural rank constraints.
The models were trained for 100 epochs on ImageNet with the LARS~\citep{you2017lars,goyal2017lars} optimizer. DINO was also trained using multi-crop.

In order to evaluate the methods, we use ImageNet (our source dataset), as well as iNaturalist18~\citep{vanhorni2018naturalist}, Places205~\citep{zhou2014places}, EuroSat~\citep{helber2019eurosat}, SUN397~\citep{xiao2010sun}, and StanfordCars~\citep{krause20133d} to evaluate the trained models on unseen datasets. While we focus on these datasets for our visualizations, we also include CIFAR10, CIFAR100~\cite{cifar}, Food101~\cite{food101}, VOC07~\cite{voc07} and CLVR-count~\cite{johnson2017clevr} for our hyperparameter selection results, and provide matching visualizations in \cref{sec:sup-datasets}. These commonly used datasets provide a wide range of scenarios that differ from ImageNet and provide meaningful ways to test the robustness of \name{}. For example, iNaturalist18 consists of 8142 classes focused on fauna and flora which requires more granularity than similar classes on ImageNet, SUN397 focuses on scene understanding, deviating from the single object and object-centric images of ImageNet, and EuroSat consists of satellite images which again differ from ImageNet. Datasets such as iNaturalist can also allow theoretical limitations to manifest themselves more clearly due to the number of classes being significantly higher than the rank of learned representations. In order to evaluate on those datasets, we rely on the VISSL library~\citep{goyal2021vissl}. We provide complete details on the pretraining and evaluation setup in \cref{sec:training-details}.

\begin{figure}[!t]
    \centering
    \includegraphics[width=\columnwidth]{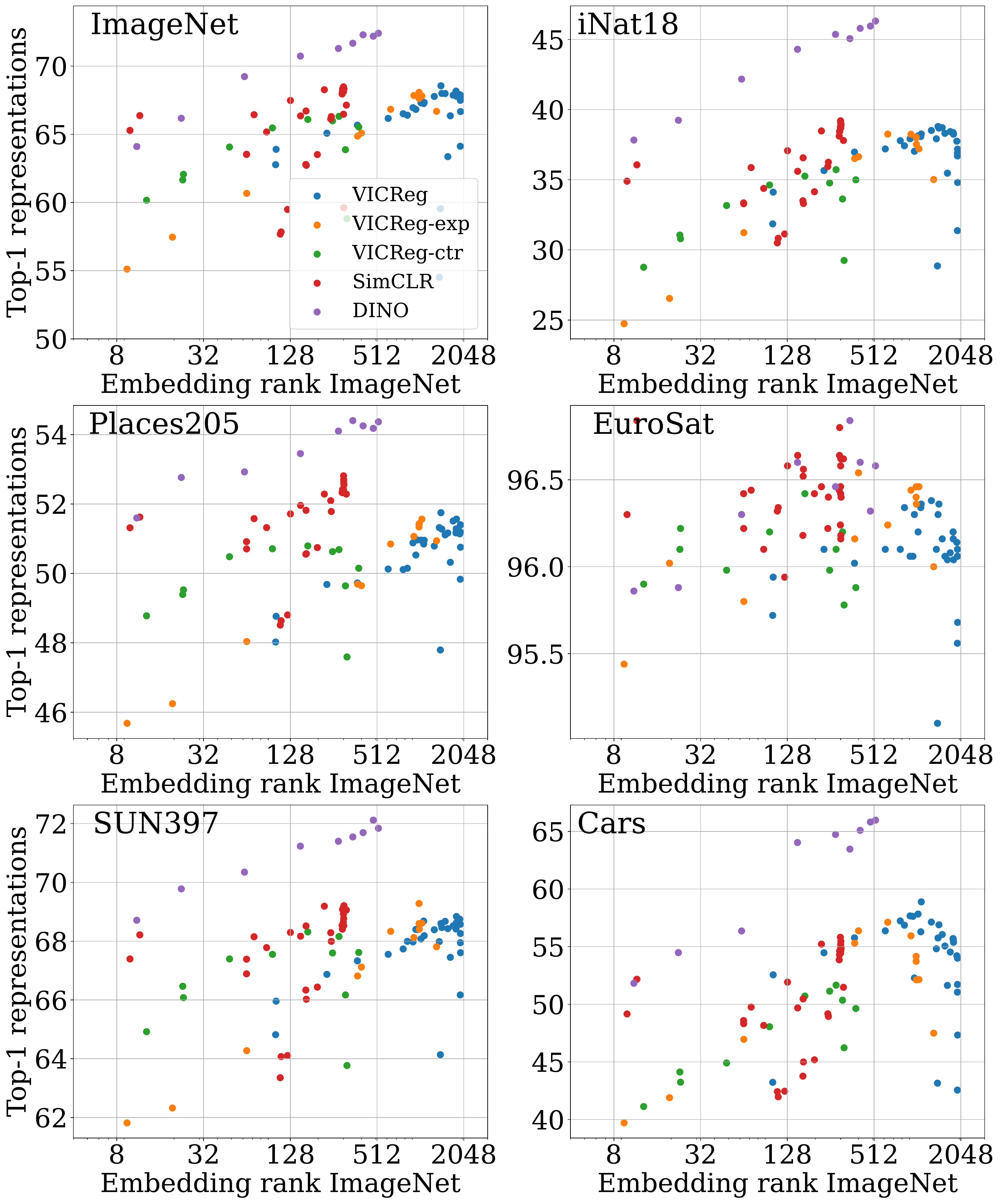}
    
    \caption{
    Validation of \name{} when evaluating performance on representations. We see that having a high rank is a necessary condition for good downstream performance.}
    \label{fig:perfs_repr}
\end{figure}

\textbf{\name{} as a prediction of linear classification accuracy.} As we can see in ~\cref{fig:perfs_repr,fig:figure_1}, for a given method the performance on the representations is improved by a higher embedding rank, whether we look on ImageNet on which the models were pretrained or on downstream datasets. This is best seen when looking at DINO, where we notice a clear trend across all datasets.
On EuroSat, the relationship is not clear since the performances are  so close between all models.
When looking at VICReg on on StanfordCars, we can clearly see that a high rank is only a necessary condition. Here the best performance is not achieved with the highest rank, even if full rank embeddings still achieve good performance. We discuss the link between rank, number of classes, and performance in \cref{sec:theory} to give some insights into \name{}'s behavior in settings with few classes such as StanfordCars.
It is also very tempting to draw conclusions when comparing different approaches, especially when looking at the ImageNet performance, however since dimensional collapse is not the only performance deciding factor one should refrain from doing so. 
\vspace{-0.2cm}
\subsection{\name{} Also Holds for Non-linear Probing}
\label{sec:nonlinear}
\label{sec:architecture}

\begin{figure}[t!] 
    \centering
    \includegraphics[width=0.48\columnwidth]{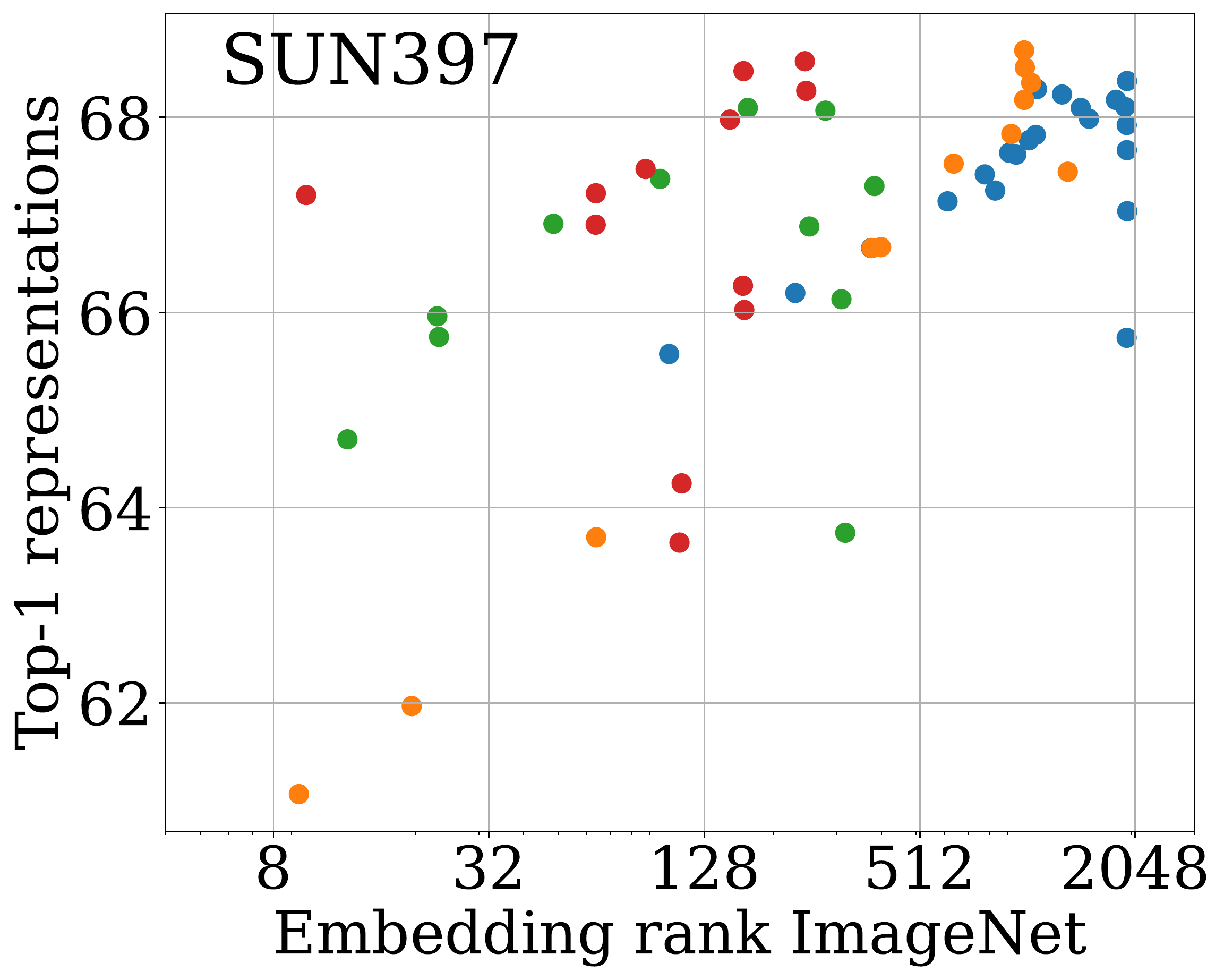}
    \hfill
    \includegraphics[width=0.48\columnwidth]{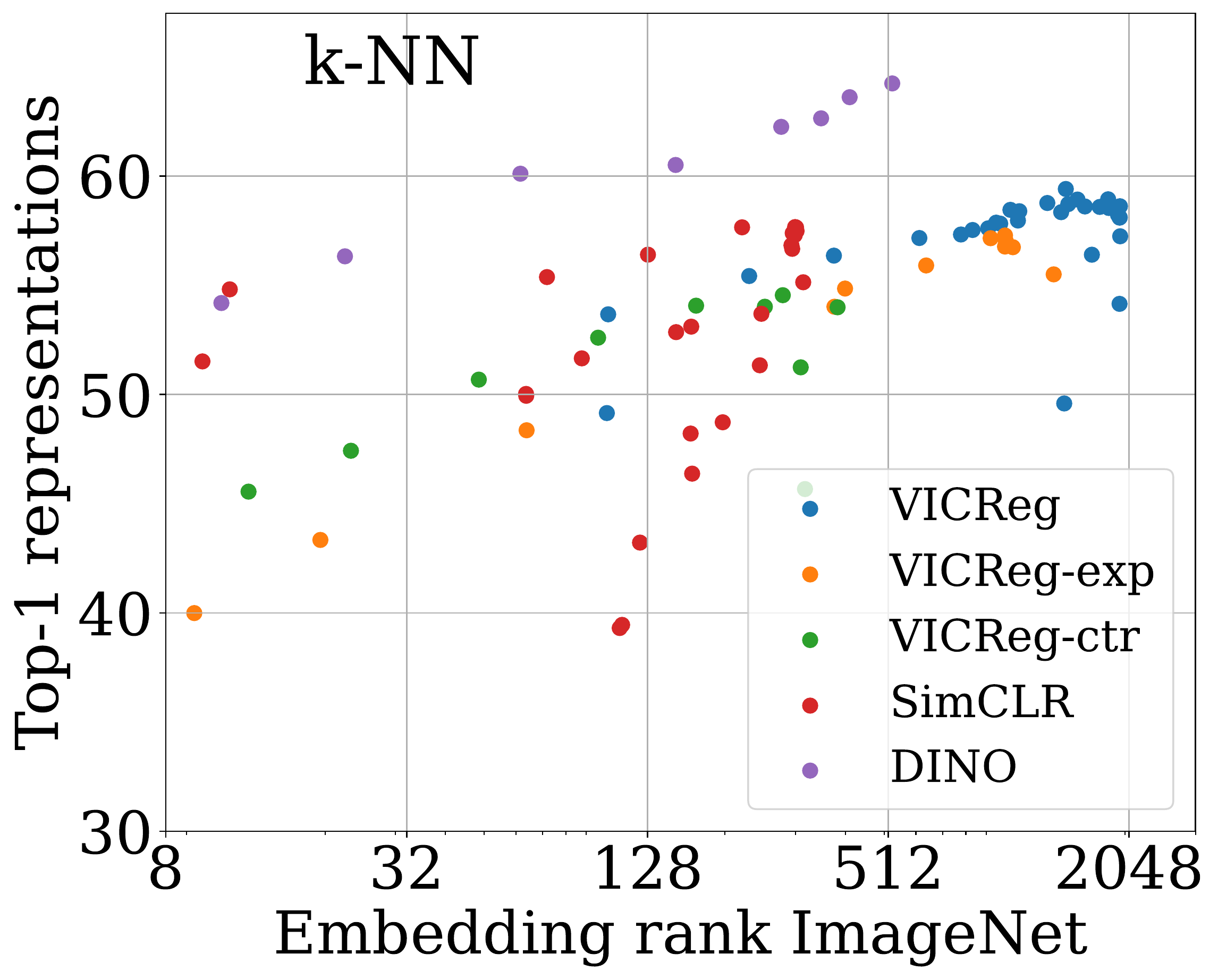}
    \vspace{-0.2cm}
    \caption{
    Impact of rank on performance on other architectures and evaluation protocols.  \textbf{(Left)} Using a 3 layer MLP as classification head does not alter the performance before or after the projector, showing that RankMe can go beyond linear evaluation. \textbf{(Right)} The same conclusion holds for k-NN evaluation on ImageNet, where RankMe remains a good indicator of performance.}    
    \label{fig:generalization}
    \vspace{-0.4cm}
\end{figure}

While we have been focusing on linear evaluation, one can wonder if the behaviors change when using a more complex task-related head. We thus give some evidence that the previously observed behaviors are similar with a non-linear classification head. we use a simple 3 layer MLP with intermediate dimensions $2048$, where each layer is followed by a ReLU activation. This choice of dimensions ensures that there are no architectural rank constraints on the embeddings. We focus on SUN397 for its conceptual difference to ImageNet. The low rank of embeddings produced by SimCLR would suggest that a non-linear classifier might help improve performance, since it is not as theoretically limited by the embeddings' rank as it is in the linear setting.
However we can see in  \cref{fig:generalization} that the behaviors for all methods are the same as in the linear regime. This would suggest that \name{} is also a suitable metric to evaluate downstream performance in a non-linear setting.
We perform the same analysis using a $k$-NN classifier, following the protocol of~\cite{zhuang2019local,caron2020swav}, where we use 36 combinations of $k$ and temperature and report the best performance. We  see in \cref{fig:generalization} that \name{} remains a good predictor of dowstream performance, with curves that are similar to what was observed with a linear classifier. Since a $k$-NN classifier evaluates the preservation of the euclidean distance instead of the linear separability, the results suggest that \name{} can extend to more evaluation protocols.

\begin{figure*}[!t]
    \begin{minipage}{0.6\textwidth}
    \centering
    \begin{algorithm}[H]
        \caption{Hyperparameter selection with \name{}}
        \begin{algorithmic}[1]
                \Require  Models $f_1,\ldots,f_N$ to compare, in increasing value of the hyperparameter
                \Require  Corresponding ranks $r_1,\ldots,r_N$
                \State $f_{best} \gets  f_1$,\; $r_{best} \gets  r_1$
                \For{ $i = 2$ to $N$}
                    \If{$ r_i > r_{best}$}
                        \State $f_{best} \gets  f_i$,\; $r_{best} \gets  r_i$
                    \ElsIf{$ r_i = r_{best}$ and ($r_i > r_{i-1}$ or $r_i > r_{i+1}$)}
                        \State $f_{best} \gets  f_i$,\; $r_{best} \gets  r_i$
                    \EndIf
                \EndFor
                \State \Return $f_{best}$
        \end{algorithmic}
    \end{algorithm}
    \end{minipage}
    \hfill
    \begin{minipage}{0.38\textwidth}
    \centering
    \includegraphics[width=\linewidth]{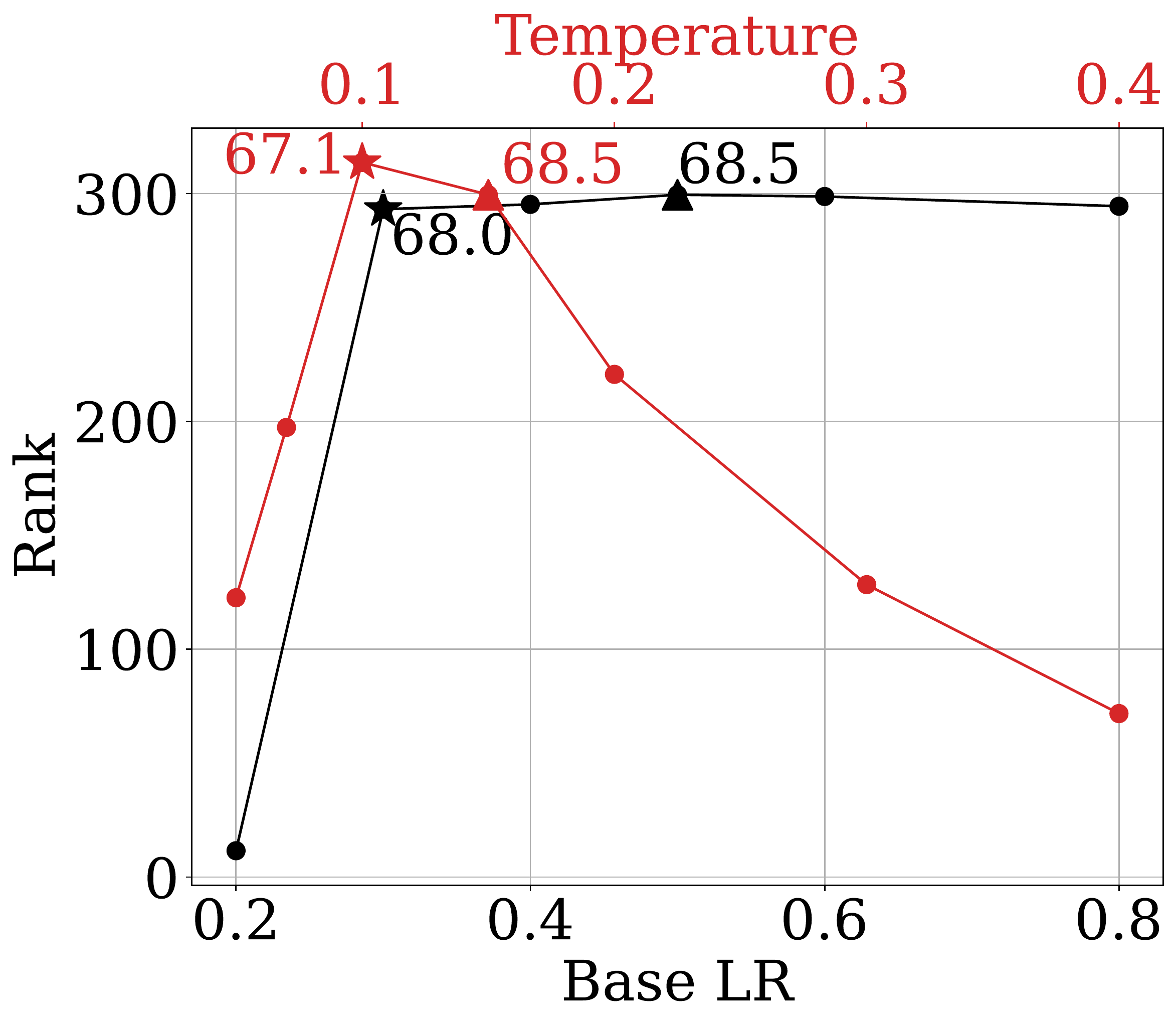}
    \end{minipage}
    
    \caption{
    {\bf(Left)} Algorithm describing how to use \name{} for hyperparameter selection. We select either the highest rank model, or if there are multiple ones, the one with the minimal/maximal value achieving it.{\bf(Right)} Visual example of the hyperparameter selection applied to SimCLR's temperature and learning rate. The star indicates the value that is selected using \name{}, and the triangle the one with the ImageNet oracle. Notice the high rank of oracle selected models.}
    \label{fig:hp-selection}
\end{figure*}

\vspace{-0.2cm}
\section{\name{} for Label-Free Cross-Validation}
\vspace{-0.2cm}

We previously focused on validating \name{} by comparing overall performance compared to linear evaluation. In this section we focus on the evolution of rank and performance when varying one hyperparameter at a time in order to demonstrate how \name{} can be used for hyperparameter selection.
We focus on loss specific hyperparameters such as the loss weights or temperature as well as hyperparameters related to optimization, such as the learning rate and weight decay. 

\vspace{-0.2cm}
\subsection{Using \name{} to select hyperparameters}

As we have shown before, having a higher rank is necessary for better performance, and using \name{} to find the best value of an hyperparameter is as simple as choosing the value that leads to the highest rank, as illustrated in ~\cref{fig:hp-selection}. Certain hyperparameters will lead to plateaus of equal rank, and for those the value that first achieves the maximal value of \name{} should be selected. This second part is however only applicable when hyperparameter values can be ordered.\\
Even in cases where the values cannot be compared, and equal ranks are found in a different setting, this still makes it possible to discard some runs and only focus on the one that achieve the maximal rank. This further highlights how maximal rank is only a necessary condition for good performance. Nonetheless, when the hyperparameters are ordered we can go one step further and use the rank alone to find a good hyperparameter value.

\vspace{-0.2cm}
\subsection{Experiments}
\vspace{-0.2cm}

\begin{table*}[!t]
  \caption{
  Top-1 accuracies obtained on the embeddings by doing hyperparameter selection using ImageNet validation performance, $\alpha$-ReQ or \name{}. OOD indicates the average performance over all the considered datasets other than ImageNet.}
  \label{tab:hp-tuning}
  \centering

  \begin{tabular}{llccccccccc}
    \toprule
     \multirow{2}{*}{Dataset} & \multirow{2}{*}{Method} &  \multicolumn{4}{c}{VICReg} &  \multicolumn{3}{c}{SimCLR} &  \multicolumn{2}{c}{DINO} \\ 
 \cmidrule(lr){3-6} \cmidrule(lr){7-9} \cmidrule(lr){9-11} &  & cov. & inv. & LR & WD & temp. & LR. & WD. & t-temp. & s-temp.\\ 
 \midrule 
\multirow{3}{*}{ImageNet} & \textcolor{gray}{ImageNet Oracle} & \textcolor{gray}{59.7} & \textcolor{gray}{59.7} & \textcolor{gray}{59.7} & \textcolor{gray}{59.7} & \textcolor{gray}{56.9} & \textcolor{gray}{56.9} & \textcolor{gray}{57.1} & \textcolor{gray}{54.6} & \textcolor{gray}{64.8}\\ 
 & $\alpha$-ReQ & \textbf{59.6} & 59.2 & 36.2 & 59.3 & 51.5 & \textbf{56.4} & 49.0 & \textbf{53.3} & 53.3\\ 
 & \name{} & \textbf{59.6} & \textbf{59.7} & \textbf{59.7} & \textbf{59.5} & \textbf{56.5} & 56.0 & \textbf{57.1} & \textbf{53.3} & \textbf{64.8}\\ 
  \midrule 
\multirow{3}{*}{OOD} & \textcolor{gray}{ImageNet Oracle} & \textcolor{gray}{55.3} & \textcolor{gray}{55.6} & \textcolor{gray}{55.3} & \textcolor{gray}{55.5} & \textcolor{gray}{54.7} & \textcolor{gray}{54.7} & \textcolor{gray}{54.7} & \textcolor{gray}{55.6} & \textcolor{gray}{60.6}\\ 
 & $\alpha$-ReQ & \textbf{55.5} & \textbf{55.7} & 48.0 & \textbf{55.1} & \textbf{56.9} & \textbf{54.6} & \textbf{54.8} &\textbf{ 52.6} & 52.6\\ 
 & \name{} & \textbf{55.5} & 55.6 & \textbf{55.3} & 55.0 & 56.4 & 54.4 & 54.7 & \textbf{52.6} & \textbf{60.6}\\ 

    \bottomrule
  \end{tabular}
\vspace{-0.2cm}
\end{table*}

In order to demonstrate the effectiveness of \name{} for hyperparameter selection, we apply the algorithm presented in ~\cref{fig:hp-selection} to find the best values for a given set of hyperparameters for VICReg, SimCLR and DINO. Our focus is on the covariance and invariance weights in VICReg, the temperature in SimCLR, on learning rate and weight decay for both, and on the student and teacher temperatures in DINO. We compare the performance on ImageNet as well as the average performance on the previously discussed OOD datasets to models selected by their ImageNet top-1 accuracy on its validation set. For per dataset performance, confer \cref{sec:hp-tables}.

\textbf{On the embeddings.}
As we can see in \cref{tab:hp-tuning}, using \name{} we are able to retrieve most of the performance on ImageNet, with gaps being lower than half a point on average. It is not possible to beat the selection using ImageNet's validation, since this is the metric we are evaluating on. However, on OOD datasets we are able to improve the performance in certain settings, while having similar performance on average. Thus, when comparing performance after the projector, \name{} is the better approach of the two to select the hyperparameters that will generalize best to unseen datasets.  When comparing to $\alpha$-ReQ, \name{} achieves better in domain performance, but on OOD datasets $\alpha$-ReQ performs slightly better, though with bigger worst case performance gaps. We provide an in-depth analysis of $\alpha$-ReQ in \cref{sec:areg}, where we find that the power law prior of $\alpha$-ReQ fails on the embeddings and as such those results must be interpreted with care.
As pointed out in~\cite{girish2022one}, using ImageNet performance to select models can lead to suboptimal performance in downstream tasks, which our results further confirm and reinforces the need for a new way of selecting hyperparameters. 

\textbf{On the representations.}
When looking at performance before the projector in ~\cref{fig:figure_1}, we can see that \name{} does not beat the models selected with ImageNet's validation set, even on OOD datasets. However, \name{} performs better than $\alpha$-ReQ in most settings, while not suffering from as severe drops in the worst cases. Nevertheless, the gaps between RankMe and the ImageNet oracle are on average of less than half a point, which shows how competitive \name{} can be for hyperparameter selection, despite using no labeled data, having no parameters to tune, and being able to be computed in a couple of minutes.

\textbf{iNat-18 pretraining.}
To show how our results extend beyond ImageNet pretraining, we applied the same protocol but pretrained our models on iNat-18. For these experiments we only compare for SimCLR's temperature and VICReg's covariance weight. Due to the high number of classes of iNat-18, we chose a projector with output dimension 8192. Since the rank cannot be higher than 2048, we apply a threshold to not choose the highest rank but the highest realistically possible. See \cref{sec:inat-pretrain} for more details. We also compare \name{} to the performance on ImageNet, to imitate a practical setting where we do not have labels for our source dataset, but have access to labels for another related one. As we can see in \cref{tab:inat-pretraining}, for VICReg's covariance weight, \name{} leads to performance similar to the iNat-18 oracle on iNat-18, but slightly outperforms it on OOD datasets. It also beats the ImageNet oracle and $\alpha$-ReQ by a significant margin. On SimCLR's temperature, we notice a small drop in performance for \name{} compared to the oracles, but it still outperforms $\alpha$-ReQ by a significant margin in all settings.
These results further reinforce the use of \name{} in general settings, even beyond ImageNet.

\begin{table}[!t]
    \centering
    \caption{Using \name{} on networks pretrained on iNat-18. We see than \name{} can improve OOD performance for VICReg, but leads to a small drop for SimCLR.}
    \label{tab:inat-pretraining}
    \begin{tabular}{llcc}
        \toprule
        Dataset & Method & Cov. & temp. \\
        \midrule
         \multirow{4}{*}{iNat-18} & \textcolor{gray}{iNat-18 Oracle} & \textcolor{gray}{36.96} & \textcolor{gray}{28.60}  \\
          & ImageNet Oracle & 35.63 & \textbf{28.60}  \\
          & $\alpha$-ReQ& 25.43 & 22.94 \\
          & \name{} & \textbf{36.89} & 27.14  \\
          \midrule
         \multirow{4}{*}{OOD} & \textcolor{gray}{iNat-18 Oracle} & \textcolor{gray}{60.70} & \textcolor{gray}{58.23}  \\
          & ImageNet Oracle & 60.65 & \textbf{58.23} \\
          & $\alpha$-ReQ& 56.51 & 56.30 \\
          & \name{} & \textbf{60.91} & 57.34 \\
          \bottomrule
    \end{tabular}
    \vspace{-0.5cm}
\end{table}

\begin{table}[!t]
    \centering
    \caption{Using \name{} on finetuning based benchmarks. The ImageNet oracle is the linear evaluation oracle. In the semi-supervised setting we report the top-1 accuracy and report the AP50 for object detection. We see that in the semi-supervised setting on ImageNet \name{} only leads to small drops in performance compared to the task or full ImageNet Oracle. For object detection we even see matching or increased performance over the ImageNet oracle.}
    \label{tab:finetuning}
    \begin{tabular}{llcc}
        \toprule
        Dataset & Method & Cov. & temp. \\
        \midrule
         \multirow{4}{*}{ImageNet-1\%} & \textcolor{gray}{Task Oracle} & \textcolor{gray}{39.7} & \textcolor{gray}{34.6}  \\
          & ImageNet Oracle & \textbf{39.7}  & \textbf{31.3}  \\
          & $\alpha$-ReQ& 39.2  & 27.3  \\
          & \name{} & 38.7 & 30.9   \\
          \midrule
          \multirow{4}{*}{ImageNet-10\%} & \textcolor{gray}{Task Oracle} & \textcolor{gray}{62.7} & \textcolor{gray}{62.6}  \\
          & ImageNet Oracle & 62.6  & \textbf{62.6}  \\
          & $\alpha$-ReQ& \textbf{62.7}  & 59.1  \\
          & \name{} & \textbf{62.7} & 61.8   \\
          \midrule
          \multirow{4}{*}{VOC07+12 (AP50)} & \textcolor{gray}{Task Oracle} & \textcolor{gray}{79.7} & \textcolor{gray}{81.8}  \\
          & ImageNet Oracle & 78.2  & \textbf{81.0}  \\
          & $\alpha$-ReQ& 79.0  & 80.3  \\
          & \name{} & \textbf{79.7} & \textbf{81.0}   \\
          \bottomrule
    \end{tabular}
    \vspace{-0.5cm}
\end{table}

\begin{figure*}[!t]
    \centering
    \includegraphics[width=1\textwidth]{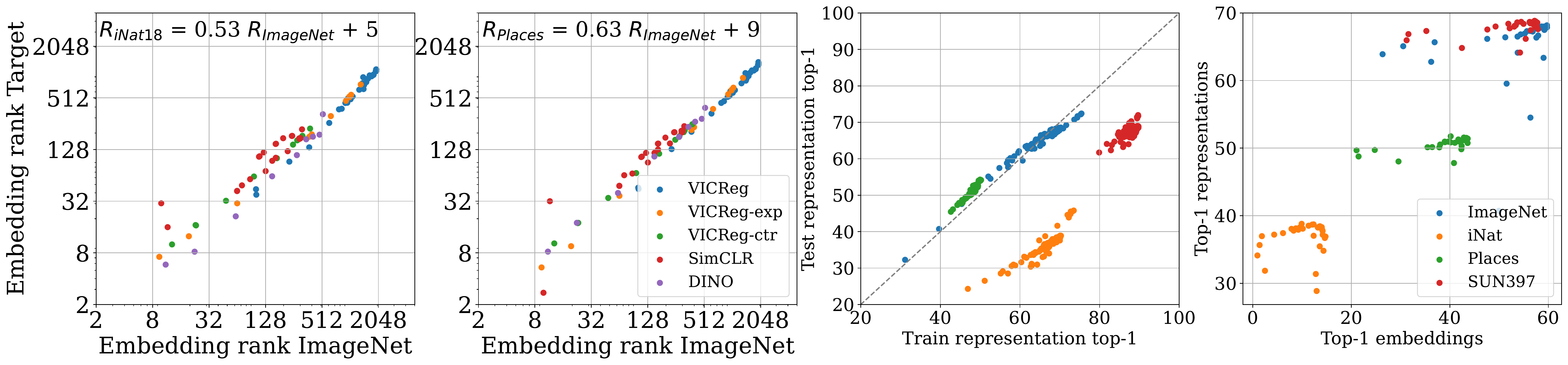}
\vspace{-0.5cm}
    \caption{
    Validation of the hypotheses motivating \name{}.\textbf{(Left, Middle Left)} Embeddings' rank transfers from source to target datasets. The estimates use $25 600$ images from the respective datasets.\textbf{(Middle Right)} Train and test accuracy are highly correlated across datasets.\textbf{(Right)} An increase in performance on embeddings leads to an increase in performance on representations.}
    \label{fig:transfer_collapse}
\end{figure*}

\textbf{Finetuning based benchmarks.}
While we have studied how \name{} is able to perform hyperparameter selection when targeting linear evaluation, finetuning based evaluations are also popular for tasks such as semi-supervised classification or object detection. Even though this setup alters the pretrained weight and thus can change the rank of the representations, our goal is to see whether \name{} can still be used when targeting these evaluations. We compare against the task oracle, $\alpha$-ReQ and to the ImageNet linear evaluation oracle, which allows us to see if linear accuracy on ImageNet is correlated for these benchmarks. We evaluate all methods on ImageNet 1\% and 10\% for semi-supervised classification, as well as on PascalVOC07+12 for object detection, following the protocol of~\cite{bardes2021vicreg}.
As we can see in \cref{tab:finetuning}, \name{} is able to retrieve most of the performance of the task oracle, except in the case of SimCLR's temperature on ImageNet-1\% where all methods lag behind the task oracle. This also shows that the linear performance on the full ImageNet dataset is not perfectly correlated with the performance in a few shot setting. There is no clear winner between $\alpha$-ReQ and \name{} in these finetuning-based evaluations, but we can see smaller drops in performance for \name{}, similarly as in previous experiments.
Nevertheless, these results suggest that even in a setting where the applicability of \name{} is not guaranteed due to the finetuning, it can still be a good method to select hyperparameters in an unsupervised fashion.

\section{\name{}: From Theory to Implementation}
\label{sec:theory}

Our goal is to build a theoretically grounded intuition into the construction of \name{}. To that hand, we first quantify approximation and classification errors of learned embeddings as a function of their rank, and then motivate how embeddings' rank can be sufficient to compare test performance of JE-SSL models's representations.

{\bf From Source Embeddings's Rank to Target Representations performance.}
We first build some intuition in the regression settings. In this case, the Eckart-Young-Mirsky theorem~\citep{eckart1936approximation} ties the best-case and worst-case approximation error of any target matrix $\mY\in\mathbb{R}^{N \times C}$ from a rank-$R$ matrix $\mP\in\mathbb{R}^{N \times C}$ to the singular values of $\mY$ that run from $R$ to the rank of $\mY$ when ordered in decreasing order. Without loss of generality, we only consider the case $C > N$ in this study, i.e., we have more samples than dimensions.
Formally, this provides a lower bound on
\begin{align*} 
\|\mY-\mP \|_F^2
\geq
\sum_{r=R+1}^{C}\sigma^2_{r}(\mY)
,
\end{align*}
which is tight for $\mP$ of rank $R$, and with $\sigma_k$ the operator returning the $k^{\rm th}$ singular value of its argument, ordered in decreasing order. This result, on which \name{} relies on, demonstrates that a necessary (but not sufficient) condition for an approximation $\mP$ to well approximate $\mY$ is to have at least the same rank as $\mY$. A similar result can be obtained in classification by considering multiple one-vs-all classifiers. In practice, however, we commonly employ a linear probe network on top of given embeddings $\mZ$ to best adapt them to the target $\mY$, i.e., $\mP=\mZ\mW+\mathbf{1}\vb^T$. However, a linear transformation is not able to increase the rank of the input matrix since
    \begin{align*}
     \rank (\mP)\leq \min(\rank(\mZ),\rank(\mW))+1.
    \end{align*}
    We directly obtain that $\min_{\mW,\vb}\|\mY-\mZ\mW-\mathbf{1}\vb^T \|_F^2
\geq
\sum_{r=R+1}^{C}\sigma^2_{r}(\mY)$. In short, the approximation lower bound is not improved by allowing linear transformation of the embeddings. Further supporting the above, we ought to recall Cover's theorem \citep{cover1965geometrical} stating that the probability of a randomly labeled set of points being linearly separable only increases if $N$ is reduced or $R$ is increased. We combine those results below.

\begin{proposition}
\label{prop:train}
     The maximum training accuracy of given embeddings in linear regression or classification increases with their rank. For classification, it plateaus when the rank surpasses the number of classes.
\end{proposition}
\vspace{-0.2cm}
By noticing that \name{} provides a smooth measure of the embeddings' rank we can lean on \cref{prop:train} to see that given two models, the one with greater \name{} value will have greater training performance. This is only guaranteed for different models of the same method, since embedding rank is not necessarily the only factor that affects performance.\\
The above result is however not too practical yet since what we are truly interested in are (i) performance on unseen samples, i.e., on the test set and out-of-distribution tasks, and (ii) performance on the representations and not the embeddings since it is common to ablate the projector network of JE-SSL models. Below, we validate three key hypotheses which, when verified, imply that we can extend the impact of \name{} such that {\em (OOD) test performance of JE-SSL representations are increased when \name{}'s value on their train set embeddings is increased}.

{\bf Validating \name{}'s Hypotheses.}
The development of \name{} is theoretically grounded when it comes to guaranteeing improved source dataset embeddings performance. To empirically extend it to target dataset representations performance we need to verify three hypotheses: (i) linear probes do not overfit, (ii) embeddings and representations performance are monotonically linked, and (iii)  source and (OOD) target embeddings ranks are monotonically linked.
Due to the different nature of datasets used for downstream tasks, there is no inherent reason for the rank of embeddings to transfer in a monotonic way to them. However, if the source dataset is diverse enough and target datasets have some semantic overlap with the source dataset, then we have
\begin{equation}
\rank(\mZ_{\rm target})\propto \rank(\mZ_{\rm source}).
\end{equation}
We observe in ~\cref{sec:ood_rank} and ~\cref{fig:transfer_collapse} that the rank of JE-SSL representations scales linearly between different input distributions e.g. going from a {\em source} task such as Imagenet~\citep{deng2009imagenet} to a {\em target} task such as iNaturalist. This is further confirmed by Pearson correlation coefficients greater than $0.99$. Interestingly, we observe that the StanfordCars dataset suffers from a less distinctive linear scaling due to the dataset distribution having a small overlap with ImageNet. This indicates that as long as the source dataset is relatively diverse, then using \name{} to select a model with greater embeddings' rank  will also correspond to selecting a model with greater embeddings' rank  on the target dataset.\\
Furthermore, as the train performance increases, so does the test performance. We validate this in the middle right of ~\cref{fig:transfer_collapse}. As a result, using \name{} to select a model with greater train performance is enough to also select a model with greater test performance.\\
Finally, we report on the right of ~\cref{fig:transfer_collapse} that the performance on embeddings and representations scales almost monotonically. These results are supported by visualizations of embeddings and representations from feature inversion models \citep{bordes2021high}. Hence, using \name{} to select the model maximizing the performance on the former also selects a model maximizing performance on the latter.\\
With these hypotheses validated empirically, we can confidently say that \name{} computed on the embeddings of the source dataset is a predictor of representations' performance on target datasets, reinforcing our experimental insights.

\vspace{-0.2cm}
\section{Conclusion}
\vspace{-0.2cm}

We have shown how the phenomenon of dimensional collapse in self-supervised learning can be used as a powerful metric to evaluate models. By using a theoretically motivated analogue of the rank of embeddings, we show that the performance on downstream datasets can easily be assessed by only looking at the training dataset, without any labels, training, or parameters. While our work focuses on linear classification, we show promising results in non-linear classification that raise the question of how general this simple metric can be. Furthermore, its competitiveness with traditional oracle based hyperparameter selection methods makes it a promising tool in settings where labels are scarce, such as in the case of large uncurated datasets. As such, this work makes a step towards completely label-less self-supervised learning, as most existing approaches' hyperparameters are tuned with the help of ImageNet's validation set. Further work will explore the use of \name{} in more varied scenarios, to further legitimize its use in designing better self-supervised approaches.

\section{Acknowledgments}
The authors wish to thank Li Jing, Gr\'{e}goire Mialon, Adrien Bardes, and Yubei Chen in no particular order, for insightful discussions. We also thank Florian Bordes for the efficient implementations that were used for our experiments.

\bibliography{references}

\begin{thebibliography}{60}
\providecommand{\natexlab}[1]{#1}
\providecommand{\url}[1]{\texttt{#1}}
\expandafter\ifx\csname urlstyle\endcsname\relax
  \providecommand{\doi}[1]{doi: #1}\else
  \providecommand{\doi}{doi: \begingroup \urlstyle{rm}\Url}\fi

\bibitem[Balestriero \& LeCun(2022)Balestriero and
  LeCun]{balestriero2022spectral}
Balestriero, R. and LeCun, Y.
\newblock Contrastive and non-contrastive self-supervised learning recover
  global and local spectral embedding methods.
\newblock \emph{arXiv preprint arXiv:2205.11508}, 2022.

\bibitem[Bardes et~al.(2021)Bardes, Ponce, and LeCun]{bardes2021vicreg}
Bardes, A., Ponce, J., and LeCun, Y.
\newblock Vicreg: Variance-invariance-covariance regularization for
  self-supervised learning.
\newblock \emph{arXiv preprint arXiv:2105.04906}, 2021.

\bibitem[Bordes et~al.(2021)Bordes, Balestriero, and Vincent]{bordes2021high}
Bordes, F., Balestriero, R., and Vincent, P.
\newblock High fidelity visualization of what your self-supervised
  representation knows about.
\newblock \emph{arXiv preprint arXiv:2112.09164}, 2021.

\bibitem[Bossard et~al.(2014)Bossard, Guillaumin, and Van~Gool]{food101}
Bossard, L., Guillaumin, M., and Van~Gool, L.
\newblock Food-101 -- mining discriminative components with random forests.
\newblock In \emph{European Conference on Computer Vision}, 2014.

\bibitem[Bromley et~al.(1994)Bromley, Guyon, LeCun, Sackinger, and
  Shah]{bromley1994siamese}
Bromley, J., Guyon, I., LeCun, Y., Sackinger, E., and Shah, R.
\newblock Signature verification using a “siamese” time delay neural
  network.
\newblock In \emph{NeurIPS}, 1994.

\bibitem[Caron et~al.(2018)Caron, Bojanowski, Joulin, and
  Douze]{caron2018clustering}
Caron, M., Bojanowski, P., Joulin, A., and Douze, M.
\newblock Deep clustering for unsupervised learning.
\newblock In \emph{ECCV}, 2018.

\bibitem[Caron et~al.(2020)Caron, Misra, Mairal, Goyal, Bojanowski, and
  Joulin]{caron2020swav}
Caron, M., Misra, I., Mairal, J., Goyal, P., Bojanowski, P., and Joulin, A.
\newblock Unsupervised learning of visual features by contrasting cluster
  assignments.
\newblock In \emph{NeurIPS}, 2020.

\bibitem[Caron et~al.(2021)Caron, Touvron, Misra, Jegou, and
  Joulin]{caron2021dino}
Caron, M., Touvron, H., Misra, I., Jegou, H., and Joulin, J. M. P. B.~A.
\newblock Emerging properties in self-supervised vision transformers.
\newblock In \emph{ICCV}, 2021.

\bibitem[Chen et~al.(2020{\natexlab{a}})Chen, Kornblith, Norouzi, and
  Hinton]{chen2020simple}
Chen, T., Kornblith, S., Norouzi, M., and Hinton, G.
\newblock A simple framework for contrastive learning of visual
  representations.
\newblock In \emph{ICML}, pp.\  1597--1607. PMLR, 2020{\natexlab{a}}.

\bibitem[Chen \& He(2020)Chen and He]{chen2020simsiam}
Chen, X. and He, K.
\newblock Exploring simple siamese representation learning.
\newblock In \emph{CVPR}, 2020.

\bibitem[Chen et~al.(2020{\natexlab{b}})Chen, Fan, Girshick, and
  He]{chen2020mocov2}
Chen, X., Fan, H., Girshick, R., and He, K.
\newblock Improved baselines with momentum contrastive learning.
\newblock \emph{arXiv preprint arXiv:2003.04297}, 2020{\natexlab{b}}.

\bibitem[Chen et~al.(2021)Chen, Xie, and He]{chen2021mocov3}
Chen, X., Xie, S., and He, K.
\newblock An empirical study of training self-supervised vision transformers.
\newblock In \emph{ICCV}, 2021.

\bibitem[Cover(1965)]{cover1965geometrical}
Cover, T.~M.
\newblock Geometrical and statistical properties of systems of linear
  inequalities with applications in pattern recognition.
\newblock \emph{IEEE transactions on electronic computers}, \penalty0
  (3):\penalty0 326--334, 1965.

\bibitem[Deng et~al.(2009)Deng, Dong, Socher, Li, Li, and
  Fei-Fei]{deng2009imagenet}
Deng, J., Dong, W., Socher, R., Li, L.-J., Li, K., and Fei-Fei, L.
\newblock Imagenet: A large-scale hierarchical image database.
\newblock In \emph{CVPR}, 2009.

\bibitem[Eckart \& Young(1936)Eckart and Young]{eckart1936approximation}
Eckart, C. and Young, G.
\newblock The approximation of one matrix by another of lower rank.
\newblock \emph{Psychometrika}, 1\penalty0 (3):\penalty0 211--218, 1936.

\bibitem[Ermolov et~al.(2021)Ermolov, Siarohin, Sangineto, and
  Sebe]{ermolov2021whitening}
Ermolov, A., Siarohin, A., Sangineto, E., and Sebe, N.
\newblock Whitening for self-supervised representation learning, 2021.

\bibitem[Everingham et~al.()Everingham, Van~Gool, Williams, Winn, and
  Zisserman]{voc07}
Everingham, M., Van~Gool, L., Williams, C. K.~I., Winn, J., and Zisserman, A.
\newblock The {PASCAL} {V}isual {O}bject {C}lasses {C}hallenge 2007 {(VOC2007)}
  {R}esults.
\newblock
  http://www.pascal-network.org/challenges/VOC/voc2007/workshop/index.html.

\bibitem[Ganea et~al.(2019)Ganea, Gelly, B{\'e}cigneul, and
  Severyn]{ganea2019breaking}
Ganea, O., Gelly, S., B{\'e}cigneul, G., and Severyn, A.
\newblock Breaking the softmax bottleneck via learnable monotonic pointwise
  non-linearities.
\newblock In \emph{International Conference on Machine Learning}, pp.\
  2073--2082. PMLR, 2019.

\bibitem[Garrido et~al.(2022)Garrido, Chen, Bardes, Najman, and
  Lecun]{garrido2022duality}
Garrido, Q., Chen, Y., Bardes, A., Najman, L., and Lecun, Y.
\newblock On the duality between contrastive and non-contrastive
  self-supervised learning.
\newblock \emph{arXiv preprint arXiv:2206.02574}, 2022.

\bibitem[Ghosh et~al.(2022)Ghosh, Mondal, Agrawal, and
  Richards]{ghosh2022investigating}
Ghosh, A., Mondal, A.~K., Agrawal, K.~K., and Richards, B.
\newblock Investigating power laws in deep representation learning.
\newblock \emph{arXiv preprint arXiv:2202.05808}, 2022.

\bibitem[Girish et~al.(2022)Girish, Dey, Joshi, Vineet, Shah, Mendes,
  Shrivastava, and Song]{girish2022one}
Girish, S., Dey, D., Joshi, N., Vineet, V., Shah, S., Mendes, C. C.~T.,
  Shrivastava, A., and Song, Y.
\newblock One network doesn't rule them all: Moving beyond handcrafted
  architectures in self-supervised learning.
\newblock \emph{arXiv preprint arXiv:2203.08130}, 2022.

\bibitem[Goyal et~al.(2017)Goyal, Dollár, Girshick, Noordhuis, Wesolowski,
  Kyrola, Tulloch, Jia, and He]{goyal2017lars}
Goyal, P., Dollár, P., Girshick, R., Noordhuis, P., Wesolowski, L., Kyrola,
  A., Tulloch, A., Jia, Y., and He, K.
\newblock Accurate, large minibatch sgd: Training imagenet in 1 hour.
\newblock \emph{arXiv preprint arXiv:1706.02677}, 2017.

\bibitem[Goyal et~al.(2021)Goyal, Duval, Reizenstein, Leavitt, Xu, Lefaudeux,
  Singh, Reis, Caron, Bojanowski, Joulin, and Misra]{goyal2021vissl}
Goyal, P., Duval, Q., Reizenstein, J., Leavitt, M., Xu, M., Lefaudeux, B.,
  Singh, M., Reis, V., Caron, M., Bojanowski, P., Joulin, A., and Misra, I.
\newblock Vissl.
\newblock \url{https://github.com/facebookresearch/vissl}, 2021.

\bibitem[Grill et~al.(2020)Grill, Strub, Altché, Tallec, Richemond,
  Buchatskaya, Doersch, Pires, Guo, Azar, Piot, Kavukcuoglu, Munos, and
  Valko]{grill2020byol}
Grill, J.-B., Strub, F., Altché, F., Tallec, C., Richemond, P.~H.,
  Buchatskaya, E., Doersch, C., Pires, B.~A., Guo, Z.~D., Azar, M.~G., Piot,
  B., Kavukcuoglu, K., Munos, R., and Valko, M.
\newblock Bootstrap your own latent: A new approach to self-supervised
  learning.
\newblock In \emph{NeurIPS}, 2020.

\bibitem[HaoChen et~al.(2021)HaoChen, Wei, Gaidon, and Ma]{haochen2021provable}
HaoChen, J.~Z., Wei, C., Gaidon, A., and Ma, T.
\newblock Provable guarantees for self-supervised deep learning with spectral
  contrastive loss.
\newblock \emph{NeurIPS}, 34, 2021.

\bibitem[He \& Ozay(2022)He and Ozay]{he2022exploring}
He, B. and Ozay, M.
\newblock Exploring the gap between collapsed \& whitened features in
  self-supervised learning.
\newblock In \emph{International Conference on Machine Learning}, pp.\
  8613--8634. PMLR, 2022.

\bibitem[He et~al.(2016)He, Zhang, Ren, and Sun]{he2016resnet}
He, K., Zhang, X., Ren, S., and Sun, J.
\newblock Deep residual learning for image recognition.
\newblock In \emph{CVPR}, 2016.

\bibitem[He et~al.(2020)He, Fan, Wu, Xie, and Girshick]{he2020moco}
He, K., Fan, H., Wu, Y., Xie, S., and Girshick, R.
\newblock Momentum contrast for unsupervised visual representation learning.
\newblock In \emph{CVPR}, 2020.

\bibitem[He et~al.(2021)He, Chen, Xie, Li, Doll{\'a}r, and Girshick]{he2021mae}
He, K., Chen, X., Xie, S., Li, Y., Doll{\'a}r, P., and Girshick, R.
\newblock Masked autoencoders are scalable vision learners.
\newblock \emph{arXiv preprint arXiv:2111.06377}, 2021.

\bibitem[He et~al.(2022)He, Chen, Xie, Li, Doll{\'a}r, and
  Girshick]{he2022masked}
He, K., Chen, X., Xie, S., Li, Y., Doll{\'a}r, P., and Girshick, R.
\newblock Masked autoencoders are scalable vision learners.
\newblock In \emph{Proceedings of the IEEE/CVF Conference on Computer Vision
  and Pattern Recognition}, pp.\  16000--16009, 2022.

\bibitem[Helber et~al.(2019)Helber, Bischke, Dengel, and
  Borth]{helber2019eurosat}
Helber, P., Bischke, B., Dengel, A., and Borth, D.
\newblock Eurosat: A novel dataset and deep learning benchmark for land use and
  land cover classification.
\newblock \emph{IEEE Journal of Selected Topics in Applied Earth Observations
  and Remote Sensing}, 12\penalty0 (7):\penalty0 2217--2226, 2019.

\bibitem[Horn et~al.(2018)Horn, Aodha, Song, Cui, Sun, Shepard, Adam, Perona,
  and Belongie]{vanhorni2018naturalist}
Horn, G.~V., Aodha, O.~M., Song, Y., Cui, Y., Sun, C., Shepard, A., Adam, H.,
  Perona, P., and Belongie, S.
\newblock The inaturalist species classification and detection dataset.
\newblock In \emph{CVPR}, 2018.

\bibitem[Hua et~al.(2021)Hua, Wang, Xue, Ren, Wang, and Zhao]{hua2021feature}
Hua, T., Wang, W., Xue, Z., Ren, S., Wang, Y., and Zhao, H.
\newblock On feature decorrelation in self-supervised learning.
\newblock In \emph{Proceedings of the IEEE/CVF International Conference on
  Computer Vision}, pp.\  9598--9608, 2021.

\bibitem[Jing et~al.(2022)Jing, Vincent, LeCun, and
  Tian]{jing2022understanding}
Jing, L., Vincent, P., LeCun, Y., and Tian, Y.
\newblock Understanding dimensional collapse in contrastive self-supervised
  learning.
\newblock In \emph{International Conference on Learning Representations}, 2022.
\newblock URL \url{https://openreview.net/forum?id=YevsQ05DEN7}.

\bibitem[Johnson et~al.(2017)Johnson, Hariharan, Van Der~Maaten, Fei-Fei,
  Lawrence~Zitnick, and Girshick]{johnson2017clevr}
Johnson, J., Hariharan, B., Van Der~Maaten, L., Fei-Fei, L., Lawrence~Zitnick,
  C., and Girshick, R.
\newblock Clevr: A diagnostic dataset for compositional language and elementary
  visual reasoning.
\newblock In \emph{Proceedings of the IEEE conference on computer vision and
  pattern recognition}, pp.\  2901--2910, 2017.

\bibitem[Krause et~al.(2013)Krause, Stark, Deng, and Fei-Fei]{krause20133d}
Krause, J., Stark, M., Deng, J., and Fei-Fei, L.
\newblock 3d object representations for fine-grained categorization.
\newblock In \emph{Proceedings of the IEEE international conference on computer
  vision workshops}, pp.\  554--561, 2013.

\bibitem[Krizhevsky et~al.(2009)Krizhevsky, Hinton, et~al.]{cifar}
Krizhevsky, A., Hinton, G., et~al.
\newblock Learning multiple layers of features from tiny images.
\newblock 2009.

\bibitem[Lee et~al.(2021)Lee, Arnab, Guadarrama, Canny, and
  Fischer]{lee2021cbyol}
Lee, K.-H., Arnab, A., Guadarrama, S., Canny, J., and Fischer, I.
\newblock Compressive visual representations.
\newblock In \emph{NeurIPS}, 2021.

\bibitem[Li et~al.(2022{\natexlab{a}})Li, Efros, and
  Pathak]{li2022understanding}
Li, A.~C., Efros, A.~A., and Pathak, D.
\newblock Understanding collapse in non-contrastive siamese representation
  learning.
\newblock In \emph{European Conference on Computer Vision}, pp.\  490--505.
  Springer, 2022{\natexlab{a}}.

\bibitem[Li et~al.(2022{\natexlab{b}})Li, Yang, Zhang, Gao, Xiao, Dai, Yuan,
  and Gao]{li2022esvit}
Li, C., Yang, J., Zhang, P., Gao, M., Xiao, B., Dai, X., Yuan, L., and Gao, J.
\newblock Efficient self-supervised vision transformers for representation
  learning.
\newblock In \emph{ICLR}, 2022{\natexlab{b}}.

\bibitem[Li et~al.(2022{\natexlab{c}})Li, Chen, LeCun, and
  Sommer]{li2022neural}
Li, Z., Chen, Y., LeCun, Y., and Sommer, F.~T.
\newblock Neural manifold clustering and embedding.
\newblock \emph{arXiv preprint arXiv:2201.10000}, 2022{\natexlab{c}}.

\bibitem[Loshchilov \& Hutter(2017)Loshchilov and Hutter]{loshchilov2017adamw}
Loshchilov, I. and Hutter, F.
\newblock Decoupled weight decay regularization.
\newblock \emph{arXiv preprint arXiv:1711.05101}, 2017.

\bibitem[Misra \& Maaten(2020)Misra and Maaten]{misra2020pirl}
Misra, I. and Maaten, L. v.~d.
\newblock Self-supervised learning of pretext-invariant representations.
\newblock In \emph{CVPR}, 2020.

\bibitem[Oord et~al.(2018)Oord, Li, and Vinyals]{oord2018infonce}
Oord, A. v.~d., Li, Y., and Vinyals, O.
\newblock Representation learning with contrastive predictive coding.
\newblock \emph{arXiv preprint arXiv:1807.03748}, 2018.

\bibitem[Press et~al.(2007)Press, Teukolsky, Vetterling, and
  Flannery]{press2007numerical}
Press, W.~H., Teukolsky, S.~A., Vetterling, W.~T., and Flannery, B.~P.
\newblock \emph{Numerical recipes 3rd edition: The art of scientific
  computing}.
\newblock Cambridge university press, 2007.

\bibitem[Reed et~al.(2021)Reed, Metzger, Srinivas, Darrell, and
  Keutzer]{reed2021selfaugment}
Reed, C.~J., Metzger, S., Srinivas, A., Darrell, T., and Keutzer, K.
\newblock Selfaugment: Automatic augmentation policies for self-supervised
  learning.
\newblock In \emph{Proceedings of the IEEE/CVF Conference on Computer Vision
  and Pattern Recognition}, pp.\  2674--2683, 2021.

\bibitem[Roy \& Vetterli(2007)Roy and Vetterli]{roy2007effective}
Roy, O. and Vetterli, M.
\newblock The effective rank: A measure of effective dimensionality.
\newblock In \emph{2007 15th European signal processing conference}, pp.\
  606--610. IEEE, 2007.

\bibitem[Santurkar et~al.(2018)Santurkar, Tsipras, Ilyas, and
  Madry]{santurkar2018does}
Santurkar, S., Tsipras, D., Ilyas, A., and Madry, A.
\newblock How does batch normalization help optimization?
\newblock \emph{Advances in neural information processing systems}, 31, 2018.

\bibitem[Shannon(1948)]{shannon1948mathematical}
Shannon, C.~E.
\newblock A mathematical theory of communication.
\newblock \emph{The Bell system technical journal}, 27\penalty0 (3):\penalty0
  379--423, 1948.

\bibitem[Tomasev et~al.(2022)Tomasev, Bica, McWilliams, Buesing, Pascanu,
  Blundell, and Mitrovic]{tomasev2022relicv2}
Tomasev, N., Bica, I., McWilliams, B., Buesing, L., Pascanu, R., Blundell, C.,
  and Mitrovic, J.
\newblock Pushing the limits of self-supervised resnets: Can we outperform
  supervised learning without labels on imagenet?
\newblock \emph{arXiv preprint arXiv:2201.05119}, 2022.

\bibitem[Ulyanov et~al.(2018)Ulyanov, Vedaldi, and Lempitsky]{ulyanov2018deep}
Ulyanov, D., Vedaldi, A., and Lempitsky, V.
\newblock Deep image prior.
\newblock In \emph{Proceedings of the IEEE conference on computer vision and
  pattern recognition}, pp.\  9446--9454, 2018.

\bibitem[Wu et~al.(2018)Wu, Xiong, Yu, , and Lin]{wu2018discrimination}
Wu, Z., Xiong, Y., Yu, S., , and Lin, D.
\newblock Unsupervised feature learning via non-parametric instance
  discrimination.
\newblock In \emph{CVPR}, 2018.

\bibitem[Xiao et~al.(2010)Xiao, Hays, Ehinger, Oliva, and
  Torralba]{xiao2010sun}
Xiao, J., Hays, J., Ehinger, K.~A., Oliva, A., and Torralba, A.
\newblock Sun database: Large-scale scene recognition from abbey to zoo.
\newblock In \emph{2010 IEEE computer society conference on computer vision and
  pattern recognition}, pp.\  3485--3492. IEEE, 2010.

\bibitem[Yeh et~al.(2021)Yeh, Hong, Hsu, Liu, Chen, and
  LeCun]{yeh2021decoupled}
Yeh, C.-H., Hong, C.-Y., Hsu, Y.-C., Liu, T.-L., Chen, Y., and LeCun, Y.
\newblock Decoupled contrastive learning.
\newblock \emph{arXiv preprint arXiv:2110.06848}, 2021.

\bibitem[You et~al.(2017)You, Gitman, and Ginsburg]{you2017lars}
You, Y., Gitman, I., and Ginsburg, B.
\newblock Large batch training of convolutional networks.
\newblock \emph{arXiv preprint arXiv:1708.03888}, 2017.

\bibitem[Zbontar et~al.(2021)Zbontar, Jing, Misra, LeCun, and
  Deny]{zbontar2021barlow}
Zbontar, J., Jing, L., Misra, I., LeCun, Y., and Deny, S.
\newblock Barlow twins: Self-supervised learning via redundancy reduction.
\newblock In \emph{ICML}, pp.\  12310--12320. PMLR, 2021.

\bibitem[Zhou et~al.(2014)Zhou, Lapedriza, Xiao, Torralba, and
  Oliva]{zhou2014places}
Zhou, B., Lapedriza, A., Xiao, J., Torralba, A., and Oliva, A.
\newblock Learning deep features for scene recognition using places database.
\newblock In \emph{NeurIPS}, 2014.

\bibitem[Zhou et~al.(2022{\natexlab{a}})Zhou, Wei, Wang, Shen, Xie, Yuille, and
  Kong]{zhou2022ibot}
Zhou, J., Wei, C., Wang, H., Shen, W., Xie, C., Yuille, A., and Kong, T.
\newblock ibot: Image bert pre-training with online tokenizer.
\newblock In \emph{ICLR}, 2022{\natexlab{a}}.

\bibitem[Zhou et~al.(2022{\natexlab{b}})Zhou, Zhou, Si, Yu, Ng, and
  Yan]{zhou2022mugs}
Zhou, P., Zhou, Y., Si, C., Yu, W., Ng, T.~K., and Yan, S.
\newblock Mugs: A multi-granular self-supervised learning framework.
\newblock 2022{\natexlab{b}}.

\bibitem[Zhuang et~al.(2019)Zhuang, Zhai, and Yamins]{zhuang2019local}
Zhuang, C., Zhai, A.~L., and Yamins, D.
\newblock Local aggregation for unsupervised learning of visual embeddings.
\newblock In \emph{ICCV}, 2019.

\end{thebibliography}
\bibliographystyle{icml2023}

\newpage
\appendix
\onecolumn

\renewcommand{\thefigure}{S\arabic{figure}}
  \renewcommand{\thetable}{S\arabic{table}}
\setcounter{figure}{0}
\setcounter{table}{0}
\section{Background}
\label{sec:background}

In order to make our work as self-contained as possible, we recall the loss functions of the methods we study. For conciseness, we refer to the outputs of the encoder as \textit{representations} and the outputs of the projection head as \textit{embeddings}, which we denote by $z_i \in \mathbb{R}^d$.
We first briefly recall that the SimCLR loss is given by
\begin{align*}          
            \mathcal{L}=-\sum_{(i,j)\in \sP}\frac{e^{CoSim( \vz_i,\vz_j)}}{\sum_{k=1}^{N}\1_{\{k\not = i\}}e^{CoSim(\vz_i,\vz_k)}},
\end{align*}
with $\sP$ the set of all positive pairs in the current mini-batch or dataset that comprise $N$ exemplars.

VICReg's loss is defined with three components. The variances loss $v$ acts as a norm regularizer for the dimensions, and the covariance loss aims at decorrelating dimensions in the embeddings. They are respectively defined as 
\begin{equation*}
    v(\mZ) = \frac{1}{d} \sum_{i=1}^d \max\left(0,1-\sqrt{\text{Var}(Z_{\cdot,i})}\right) \; \text{and} \;  c(\mZ) = \frac{1}{d} \sum_{i\neq j} \text{Cov}(\mZ)^2.
\end{equation*}
Both of these loss are combined with an invariance loss that matches positives pairs, giving a final loss of
\begin{equation*}
   \mathcal{L} = \lambda \sum_{(i,j)\in \sP}\|z_i - z_j \|_2^2 + \mu\; c(\mZ) + \nu\; v(\mZ).
\end{equation*}

VICReg-exp is defined similarly, but with the exponential covariance loss defined as 
\begin{equation}
c_{exp}(\mZ) = \frac{1}{d}\sum_i \log\left(\sum_{j \neq i} e^{\text{Cov}(\mZ)_{i,j}/\tau} \right).
\end{equation}
VICReg-ctr is then VICReg-exp but applied to $\mZ^T$, making it a contrastive approach and conceptually similar to SimCLR.
These methods give us different scenarios of collapse and allow us to make a more general study of the rank of representations as a powerful metric.

\section{Visualizations on iNaturalist-18 \label{sec:inat-pretrain}}

\begin{figure}[ht]
    \centering
    \includegraphics[width=0.5\textwidth]{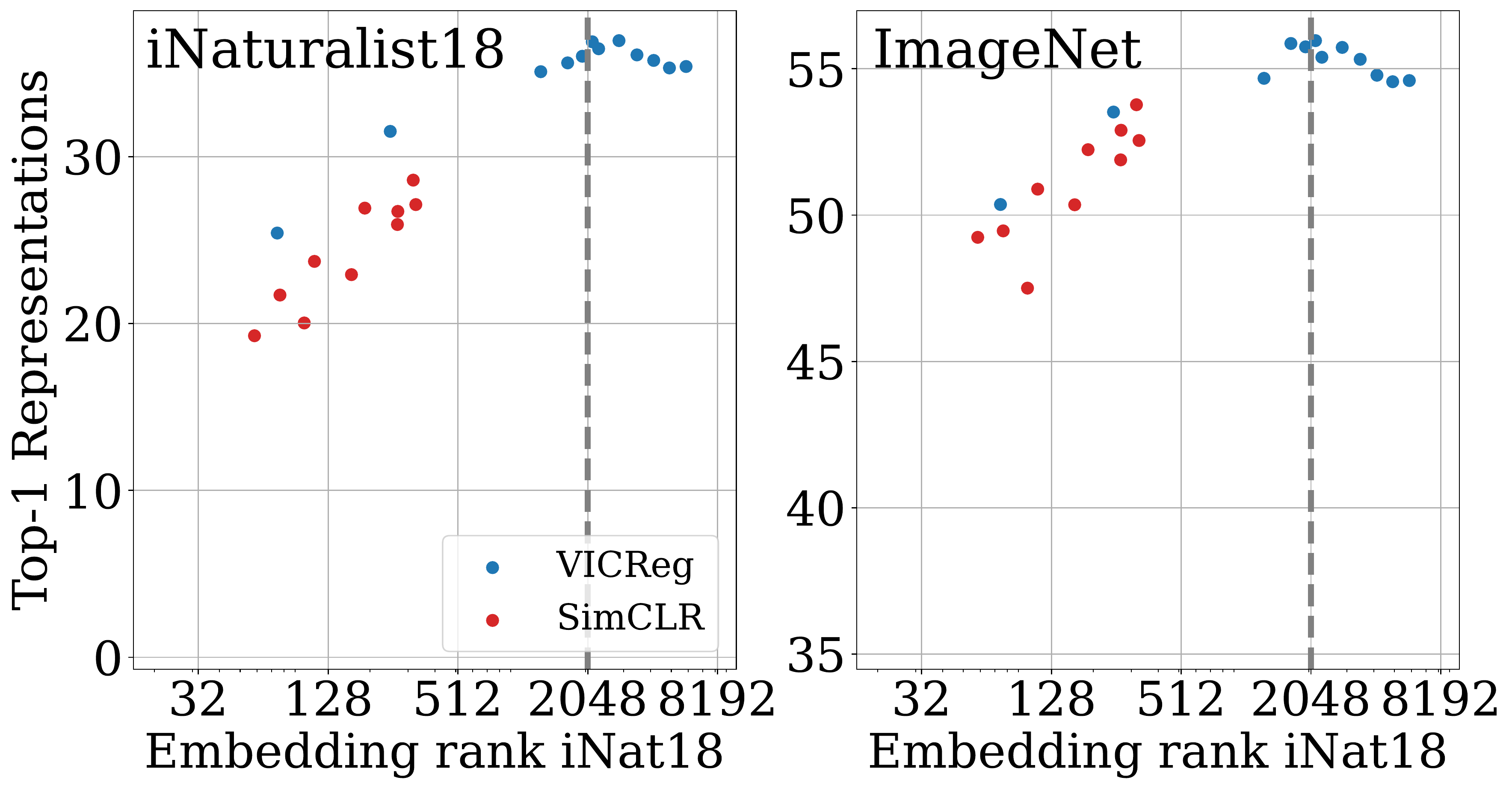}

    \caption{ RankMe applied to iNaturalist18 pretrainings. The vertical line indicates the rank constraint placed by the representation size, and so any rank above should be counted as 2048.}
    \label{fig:inat-rankme}
\end{figure}

As we can see in~\cref{fig:inat-rankme}, \name{} produces curves with the same trend as on ImageNet, for both SimCLR and VICReg. We can see that VICReg leads to ranks that go  beyond 2048, but the dimension of the manifold formed by the embeddings cannot be higher than 2048 due to the dimension of the representations. As such, for any practical purpose we clip the value of \name{} at 2048.

\section{Applicability to cluster based methods \label{sec:dino}}
While we have studied the applicability of RankMe on contrastive methods, cluster based methods such as DINO have become extremely popular, and since the definition of embeddings is not as clear cut in them, a thorough analysis is required. We will proceed in two steps:
\begin{itemize}
    \item Show that dimensional collapse happens right before the clustering layer, and on the prototypes
    \item Show that RankMe is a good measure of performance on DINO
\end{itemize}

\begin{figure}[ht]
    \centering
    \includegraphics[width=0.55\textwidth]{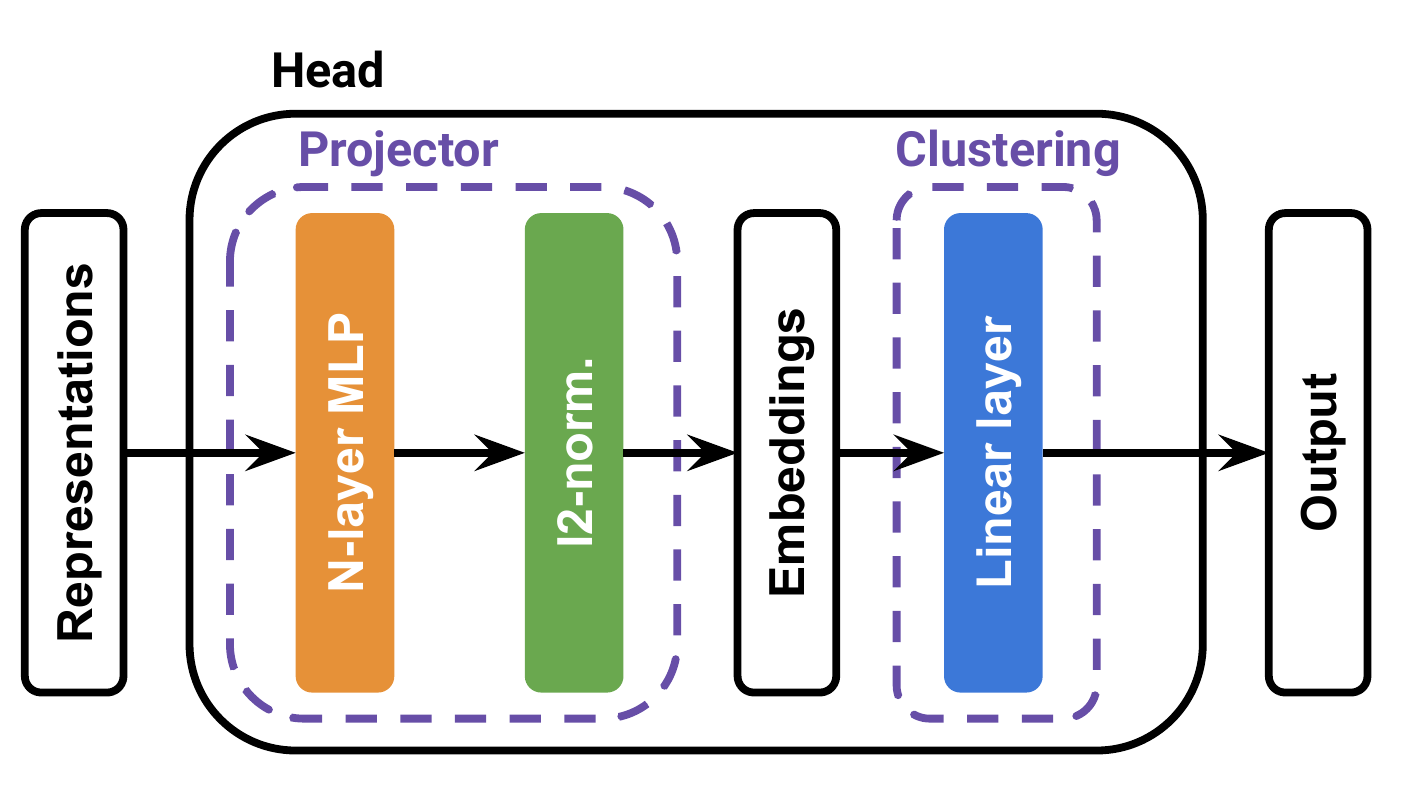}
    \hfill
    \includegraphics[width=0.4\textwidth]{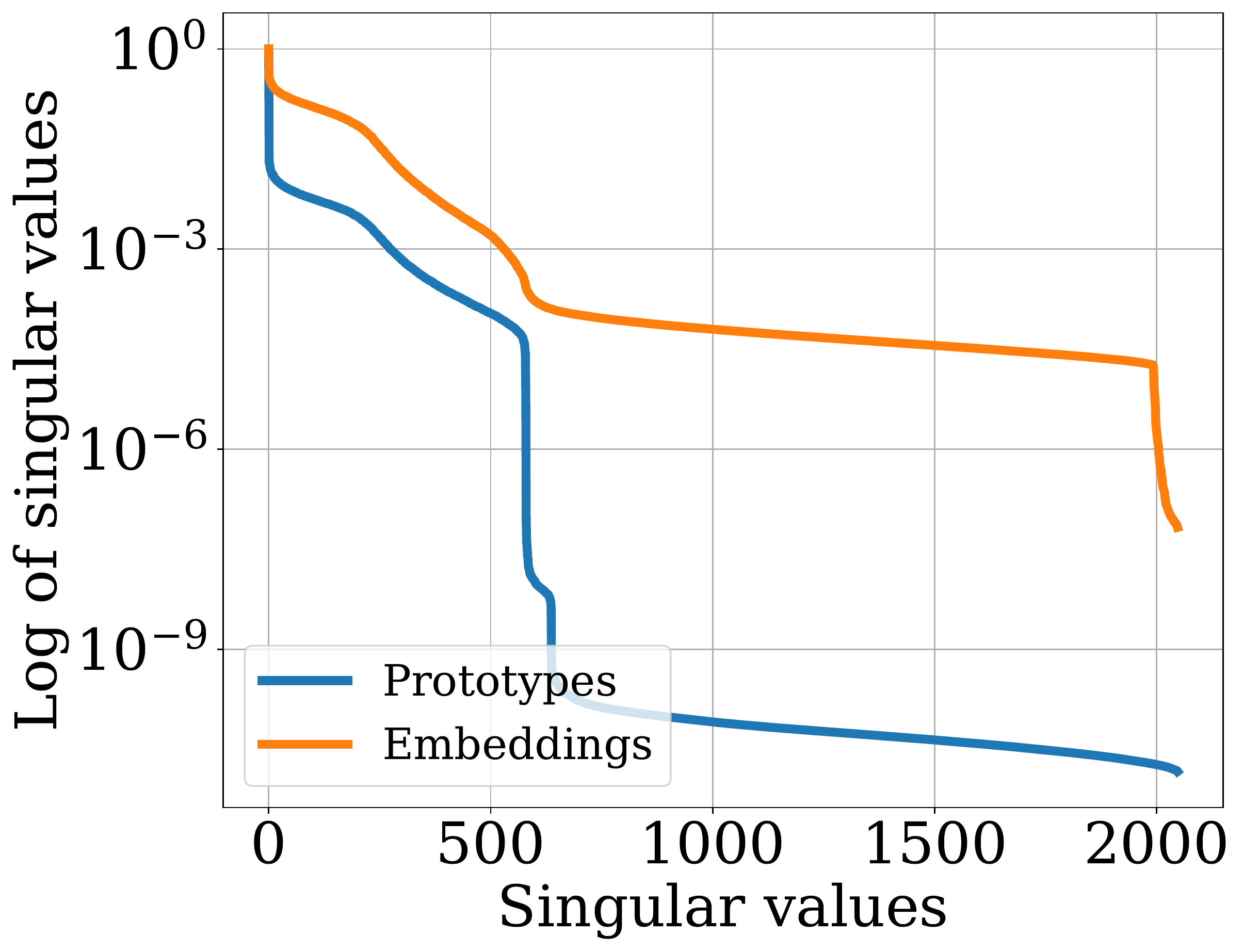}
    \caption{DINO's projection head can be split in two parts, a classical projector and a clustering layer \textbf{(Left)}. Collapse happens before the clustering layer and not on the clustering prototypes \textbf{(Right)}.}
    \label{fig:dino-collapse}
\end{figure}

As we can see in figure~\ref{fig:dino-collapse}, DINO's projector can be interpreted as both a classical projector and a clustering layer, whose weights are clustering prototypes. This interpretation comes from the softmax that is applied on the output of the projection head which can be interpreted as an InfoNCE between the embeddings and the clustering prototypes that make up the clustering layer. We see that both the embeddings and the clustering prototypes are collapsed, though at different levels.

\begin{figure}[ht]
    \begin{minipage}[c]{0.55\textwidth}
    \centering
    \includegraphics[width=\textwidth]{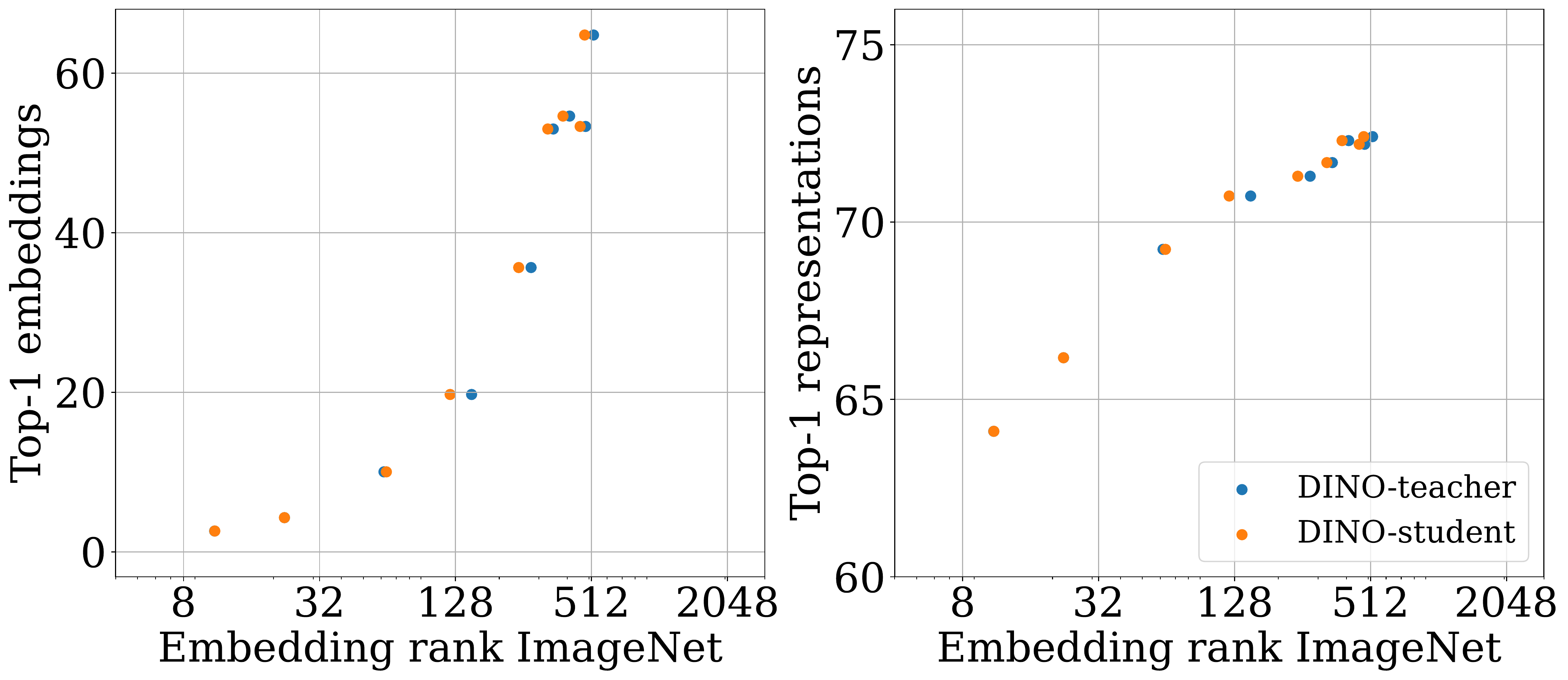}
    \end{minipage}
    \hfill
    \begin{minipage}[c]{0.44\textwidth}
  \begin{tabular}{llcc}
    \toprule
     \multirow{2}{*}{Dataset} & \multirow{2}{*}{Method} &  \multicolumn{2}{c}{DINO}\\ 
 \cmidrule(lr){3-4} & &   t-temp. & s-temp. \\ 
 \midrule 
\multirow{3}{*}{ImageNet} & \textcolor{gray}{ImageNet Oracle}& \textcolor{gray}{72.3} & \textcolor{gray}{72.4} \\ 
 & $\alpha$-ReQ & 71.7 & 66.2 \\ 
 & \name{}-embs & 72.2 & 72.4 \\ 
  & \name{}-prots & 72.3 & 72.4 \\ 

  \midrule 
\multirow{3}{*}{OOD} & ImageNet Oracle & 71.9 & 72.5\\ 
 & $\alpha$-ReQ & 71.8 & 68.5 \\ 
 & \name{}-embs & 71.8 & 72.5 \\ 
 & \name{}-prots & 71.9 & 72.5 \\ 

    \bottomrule
  \end{tabular}
    \end{minipage}

    \caption{RankMe is able to measure DINO's performance on its source dataset \textbf{(Left)}. DINO's hyperparameters can be selected by using \name{}, even by doing so directly on the prototypes \textbf{(Right)}.}
    \label{fig:dino-rankme}
\end{figure}

As we can see in \cref{fig:dino-rankme}, the phenomenon of dimensional collapse is highly visible in DINO, which enables the use \name{} to find optimal hyperparameter values. While in \cref{fig:figure_1} we applied \name{} to the embeddings to be consistent with other methods, we see that it can be applied directly to the prototypes, yielding very similar results and matching the ImageNet oracle here. The main advantage coming from using prototypes is that they are already computed during training, and as such the application of \name{} does not require computing any embeddings. This makes \name{} even more appealing for clustering based methods where such technique can be applied.

\section{Complete visualizations on all datasets \label{sec:sup-datasets}}

While we previously focused on certain datasets for their interesting natures, we provide additional visualizations for the remaining datasets, as well as for performance on the embeddings.
\begin{figure}[ht]
    \centering
    \includegraphics[width=1\textwidth]{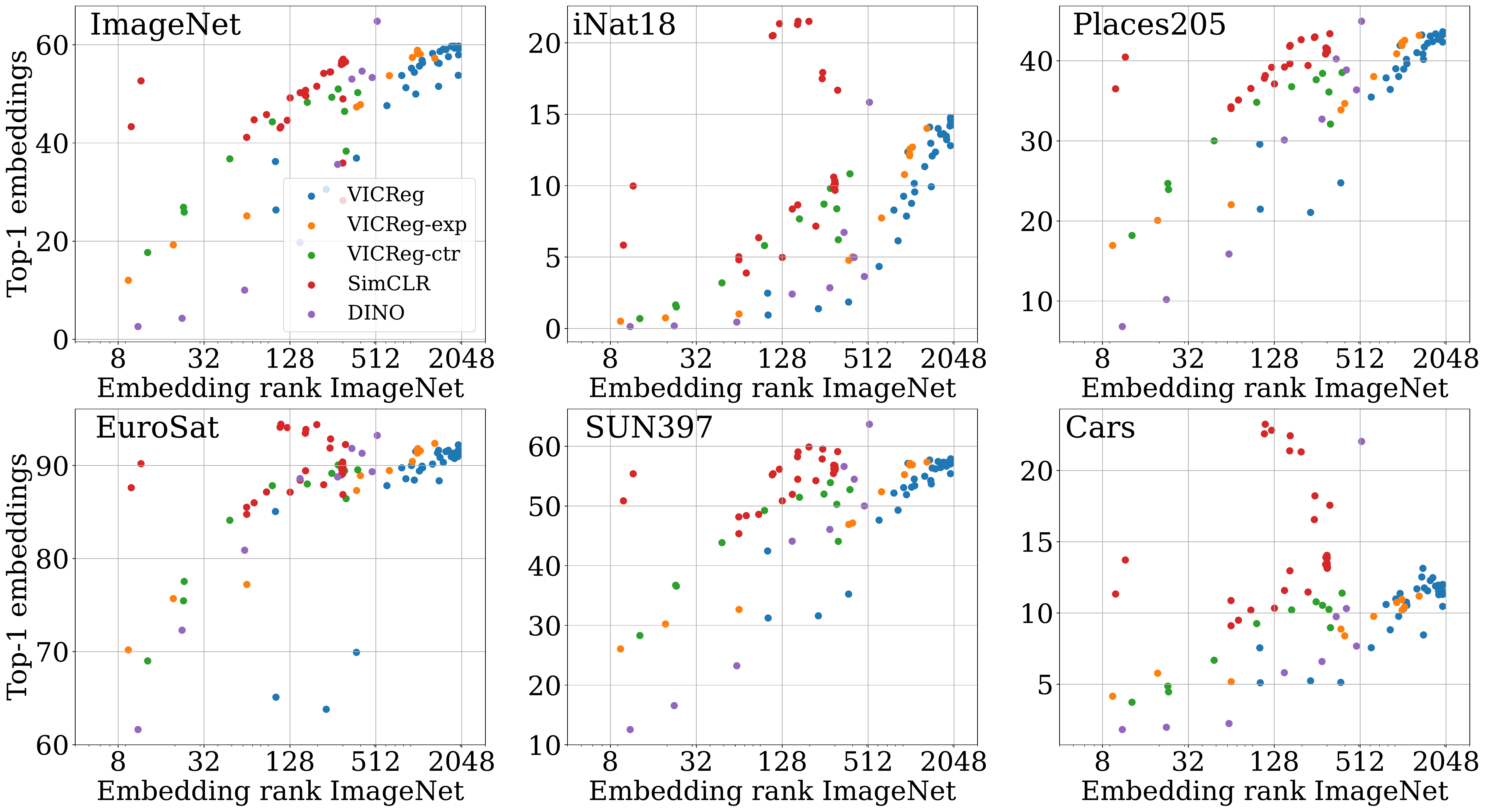}
    \includegraphics[width=1\textwidth]{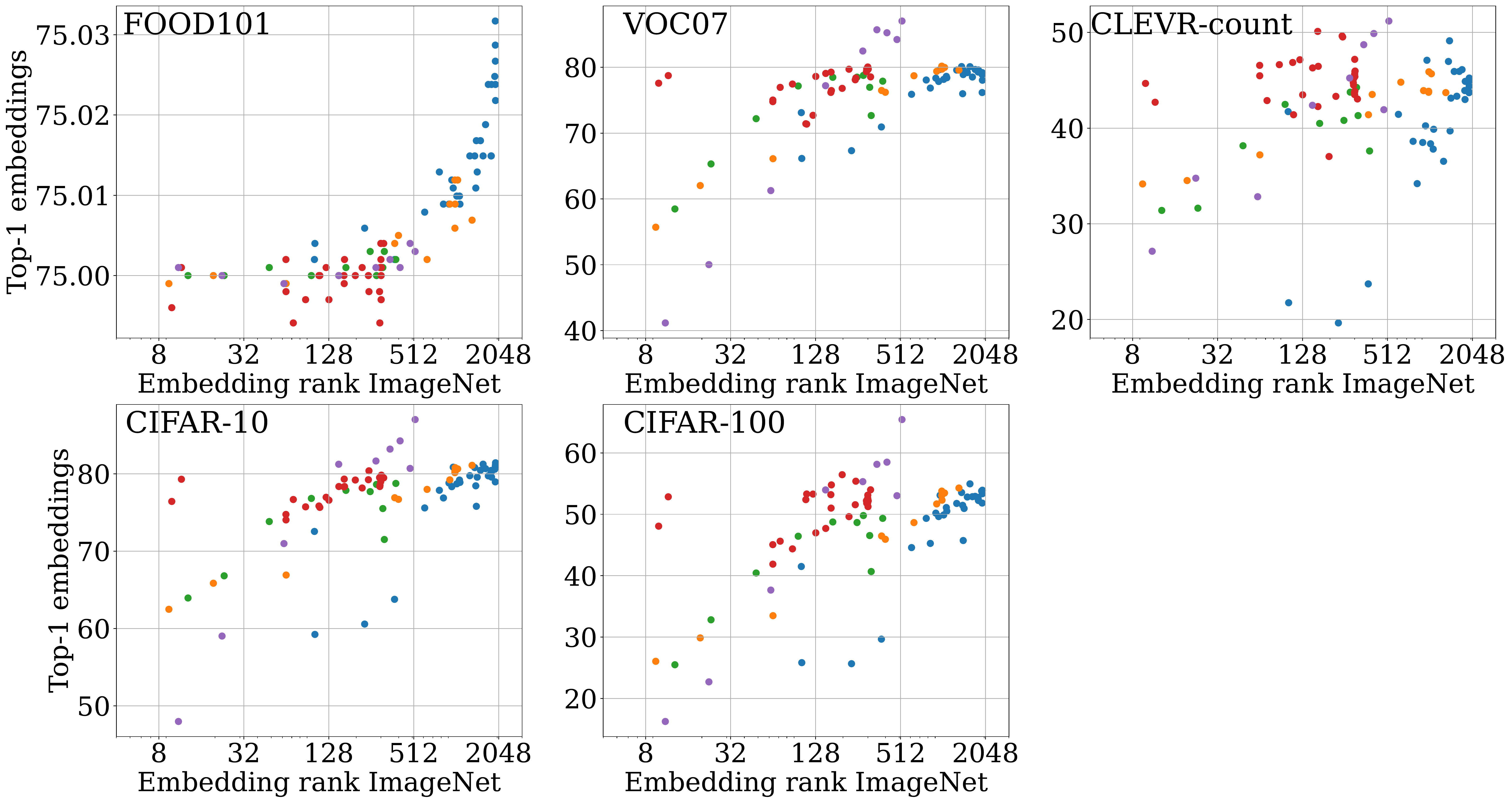}
    \caption{Link between embedding rank and downstream performance on the embeddings.}
    \label{fig:perfs-embs-supdatasets}
\end{figure}
\begin{figure}[ht]
    \centering
    \includegraphics[width=1\textwidth]{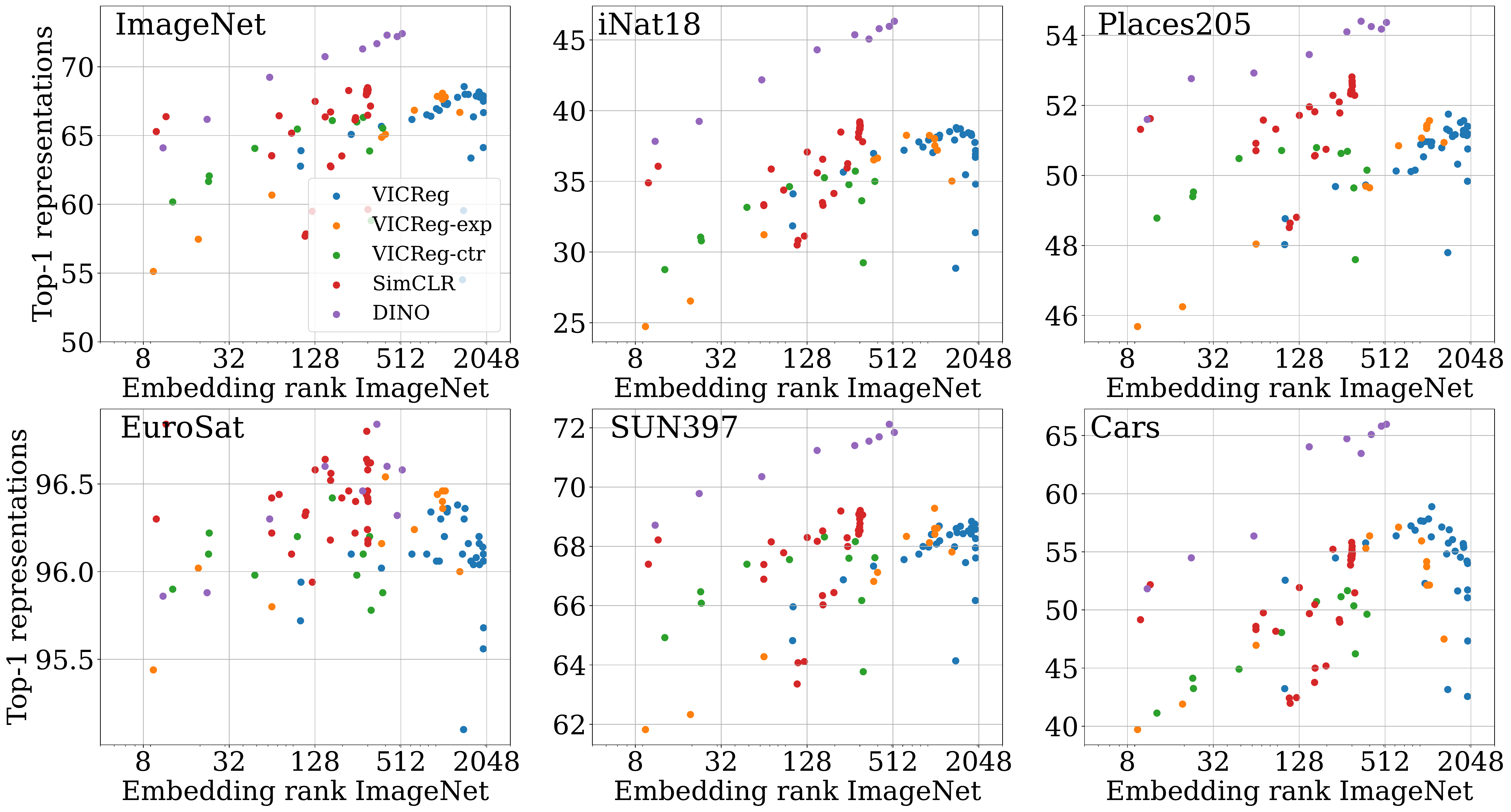}
    \includegraphics[width=1\textwidth]{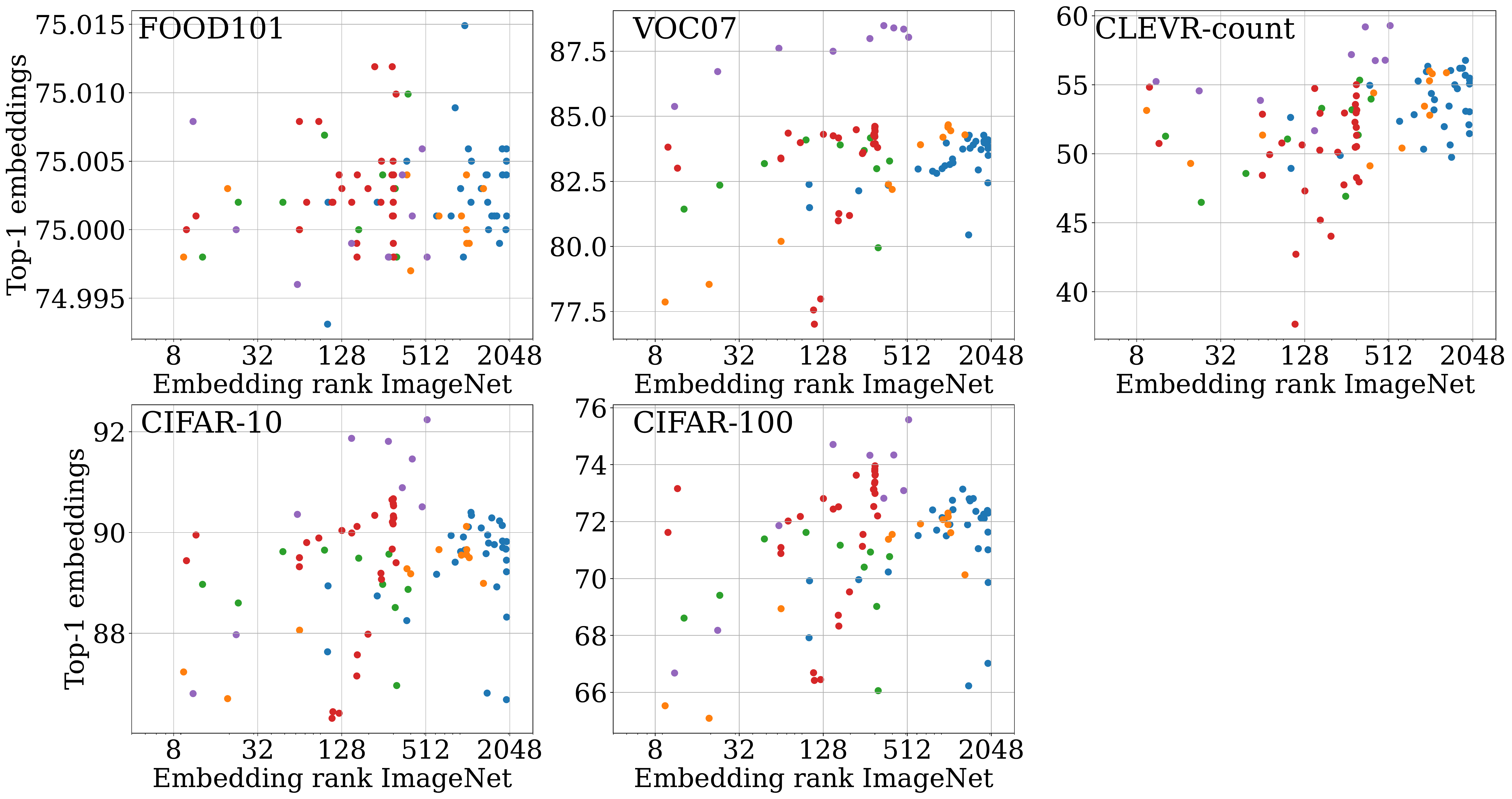}
    \caption{Link between embedding rank and downstream performance on the representations.}
    \label{fig:perfs-reprs-supdatasets}
\end{figure}

As we can see in \cref{fig:perfs-reprs-supdatasets,fig:perfs-embs-supdatasets}, we find similar behaviors as before, apart from Food101 where performance are almost identical for all methods. This reinforces the previous validation of RankMe. The relative simplicity of the datasets targeted here makes the theoretical limitations of rank-deficient embeddings harder to see, even though we still see that a high rank helps generalization.

\clearpage
\section{Detailed results for $\alpha$-ReQ \label{sec:areg}}
\begin{figure}[ht]
    \centering
    \includegraphics[width=1\textwidth]{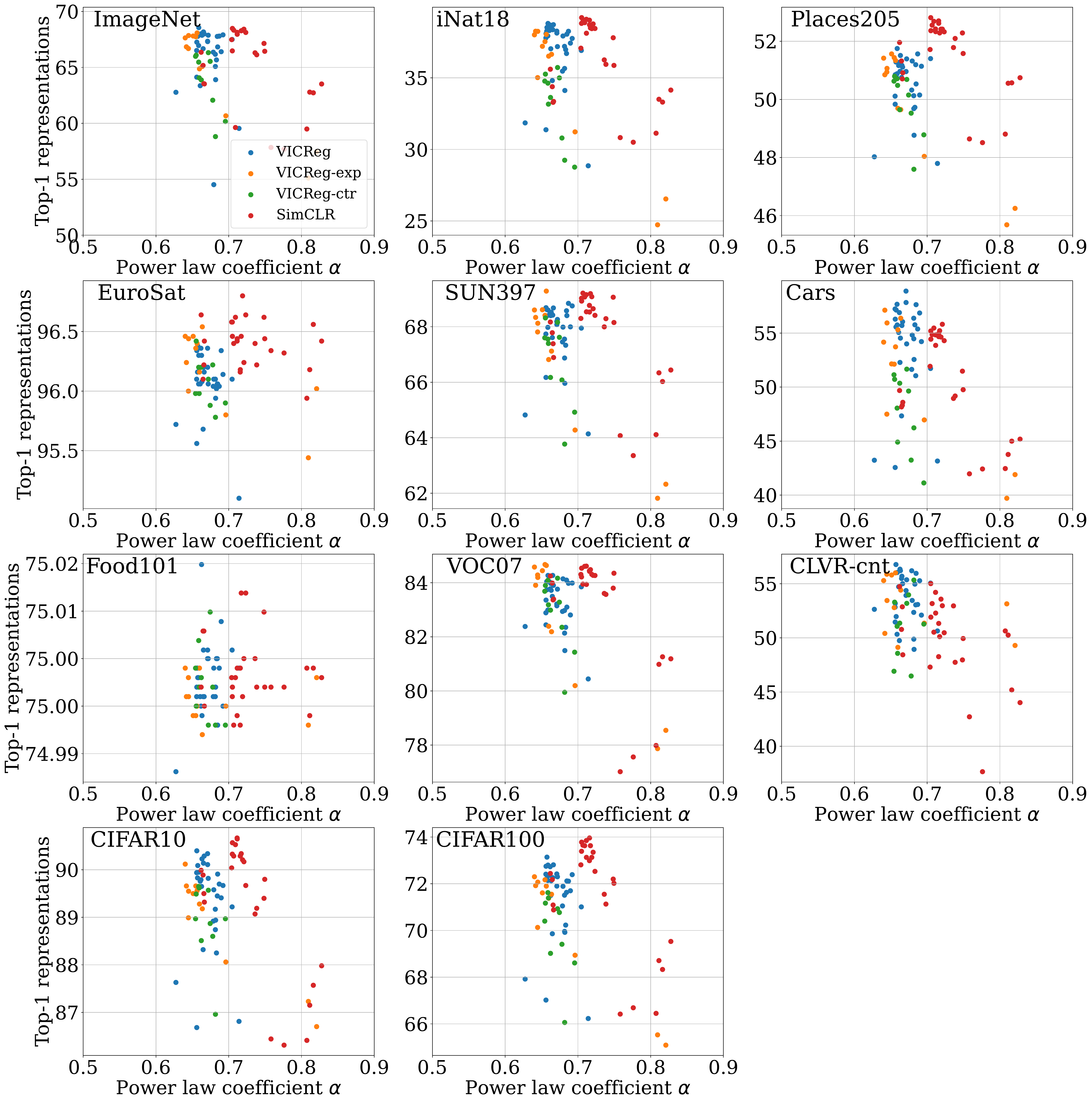}
    \caption{Link between $\alpha$-ReQ measured on the representations and performance on the representations.}
    \label{fig:areg}
\end{figure}

\begin{figure}[ht]
    \centering
    \includegraphics[width=1\textwidth]{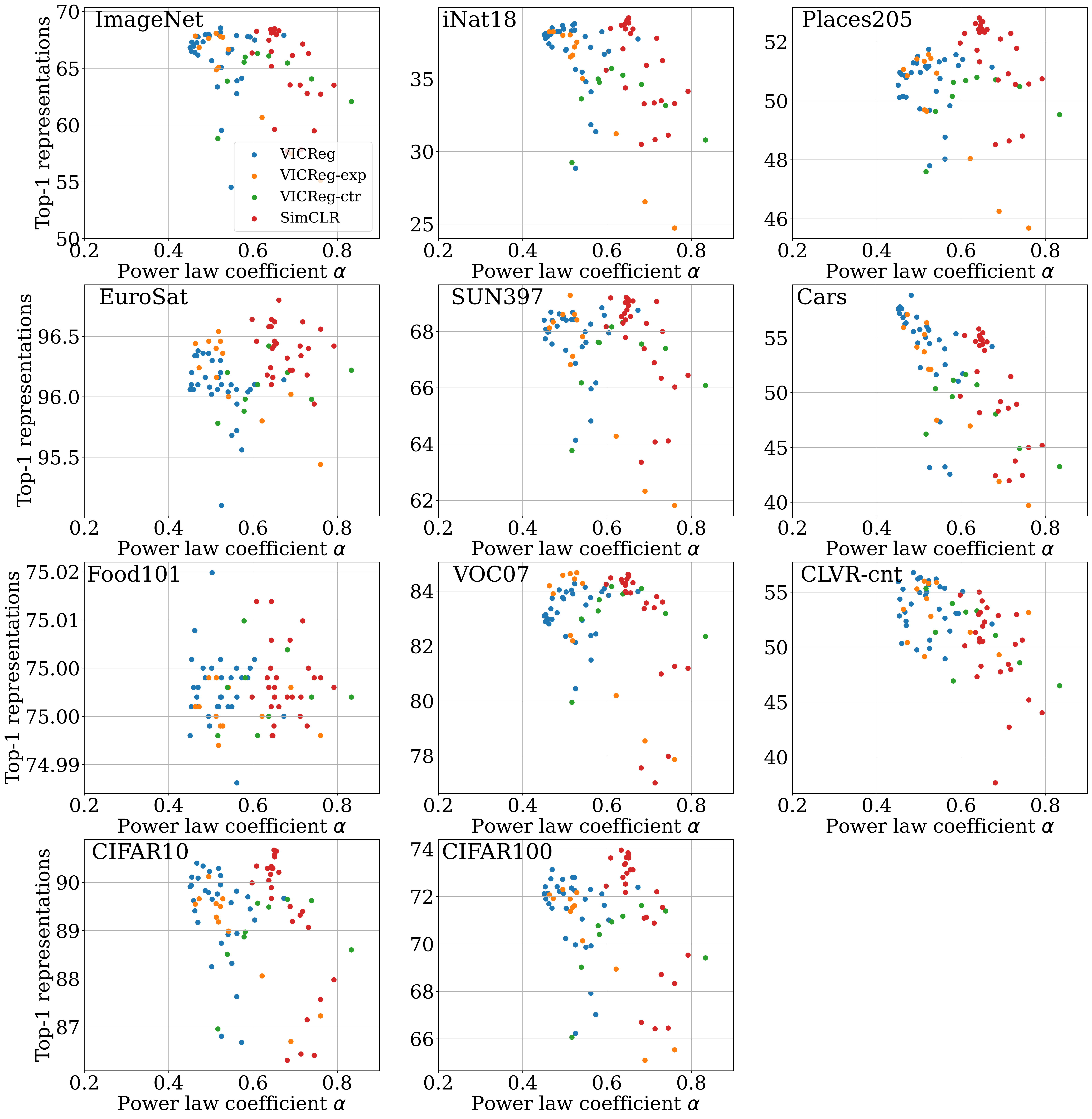}
    \caption{Link between $\alpha$-ReQ measured on the embeddings and performance on the representations.}
    \label{fig:areg-embs}
\end{figure}

In order to further study the performance of $\alpha$-ReQ, we reproduce our plots for \name{} using $\alpha$-ReQ instead of the rank of embeddings. We compare both the intended use of $\alpha$-ReQ in \cref{fig:areg}, as well as applying it on the embeddings to measure performance on the representations, which we found was necessary for \name{} in \cref{fig:areg-embs}. We do not include DINO in those plots for readability, as it would force us to change the x-axis scale, making the results harder to interpret.

As we can see in \cref{fig:areg}, there are no clear link visible between the value of $\alpha$-ReQ and downstream performance. Especially, we are unable to see the tendency of performance to increase as $\alpha$ tends to one. Nonetheless $\alpha$-ReQ was still able to lead to good performance when used for hyperparameter selection. 

When applying $\alpha$-ReQ as we would \name{}, we can see in \cref{fig:areg-embs} that there is again no trend of performance increase when $\alpha$ tends to one. On the contrary we even find that performance tends to get better with a lower $\alpha$, as is most visible on StanfordCars, iNaturalist18 or ImageNet for example. $\alpha$ going towards zero means that the singular values of the embeddings tends to a uniform distribution, in line with the goal of \name{}. 

\begin{figure}[ht]
    \centering
    \includegraphics[width=1\textwidth]{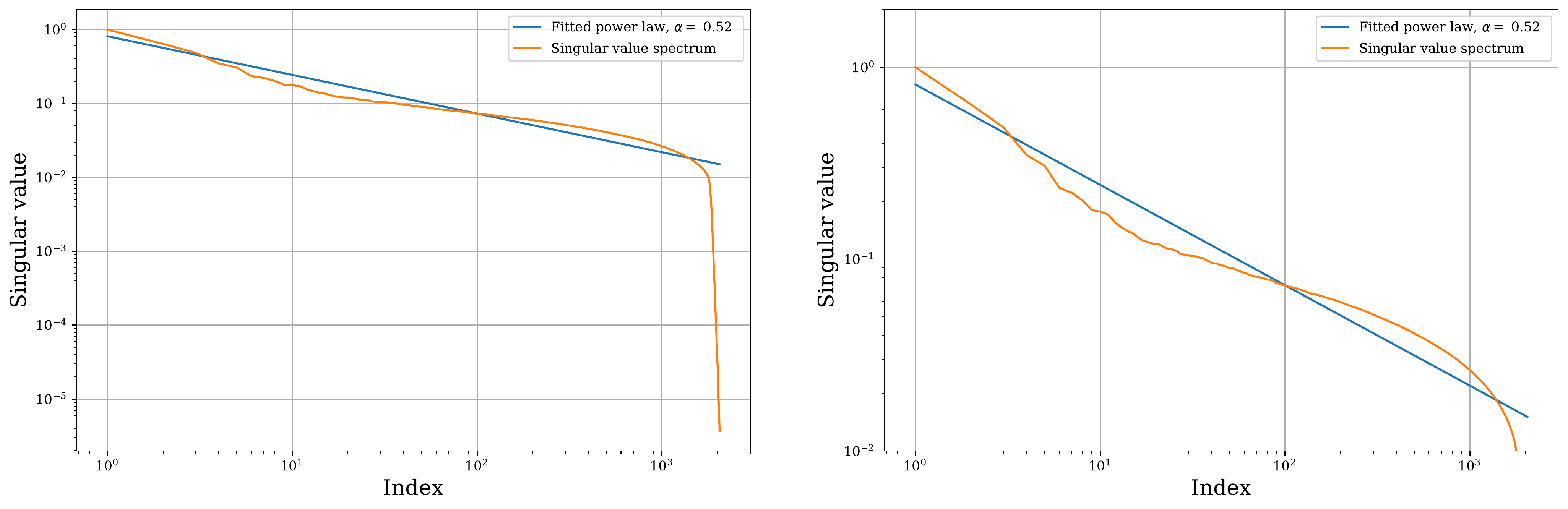}
    \caption{Validation of the power-law prior on un-collapsed representations.\textbf{(Left)} Overall visualization. \textbf{(Right)} Zoom on the high singular values. }
    \label{fig:pl-prior-vicreg}
\end{figure}

\begin{figure}[ht]
    \centering
    \includegraphics[width=1\textwidth]{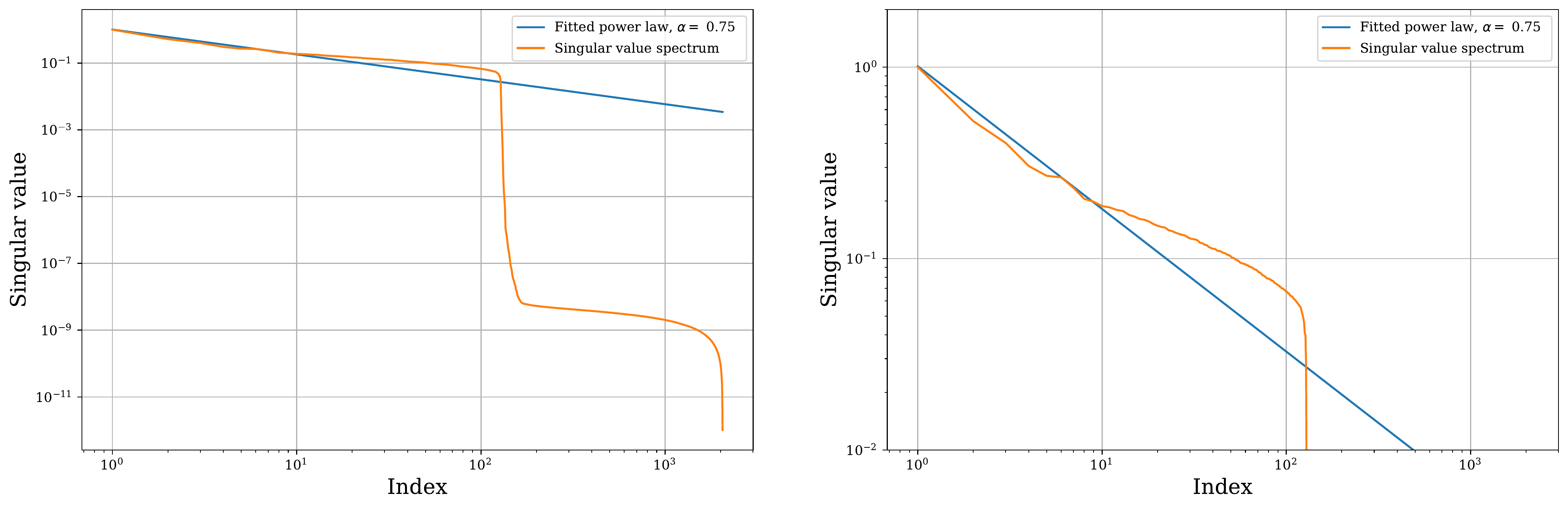}
    \caption{ The power-law prior does not hold on collapsed representations.\textbf{(Left)} Overall visualization. \textbf{(Right)} Zoom on the high singular values. }
    \label{fig:pl-prior-simclr}
\end{figure}

As we can see in \cref{fig:pl-prior-vicreg,fig:pl-prior-simclr}, the power-law prior of $\alpha$-ReQ holds well in the case of non-collapsed embeddings, but when we apply it on collapsed ones, this assumptions fails. It even provides a poor approximation of the main rank "plateau" with the highest singular values as can be seen on the right of \cref{fig:pl-prior-simclr}. This further confirms the findings of \cite{he2022exploring}, and shows that when applying $\alpha$-ReQ directly on the embeddings one must be careful since the core assumptions of the method is violated.

\clearpage
\section{Comparison of the rank estimators}\label{sec:rank-comparison}

\begin{figure}[ht]
    \centering
    \includegraphics[width=0.8\textwidth]{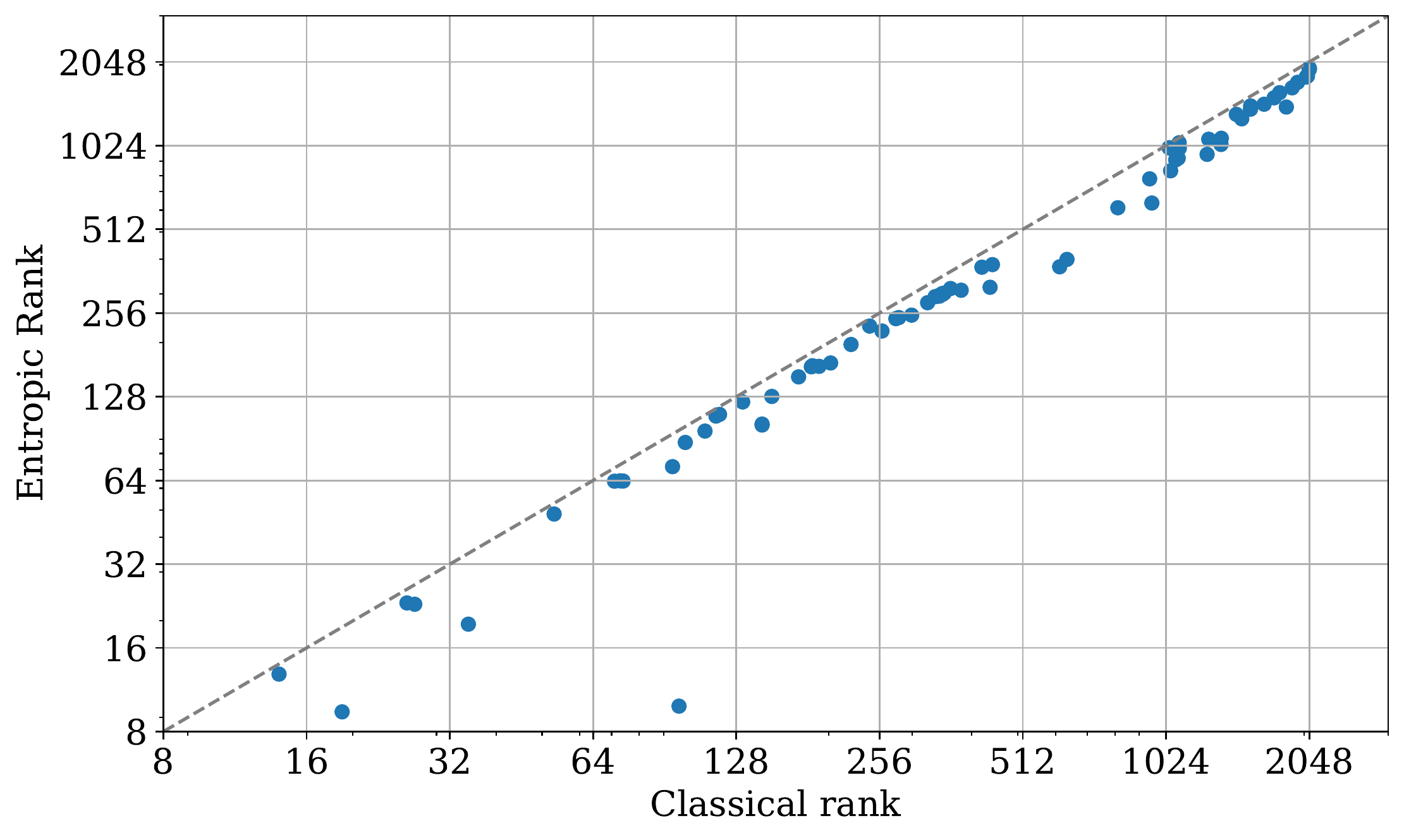}
    \caption{Relationship between the two rank estimators, Pearson correlation coefficient of $0.99$. Outliers indicate embeddings with singular values to the threshold, showing how the entropic rank takes into account this information.}
    \label{fig:ranks-correlation}
\end{figure}

Since we do not rely on the classical threshold-based rank estimator, it is important to verify how well our entropy based one correlates with it.
As we can see in \cref{fig:ranks-correlation}, both estimates discussed previously correlate extremely well, showing that using one or the other should not lead to significant differences, as validated in \cref{sec:repro-rank}. Nonetheless, the entropic estimator takes into account the degree of whitening of the embeddings, which links better to theoretical results.

\section{Convergence of the rank estimators}\label{sec:rank-convergence}

\begin{figure}[ht]
    \centering
    \includegraphics[width=0.9\textwidth]{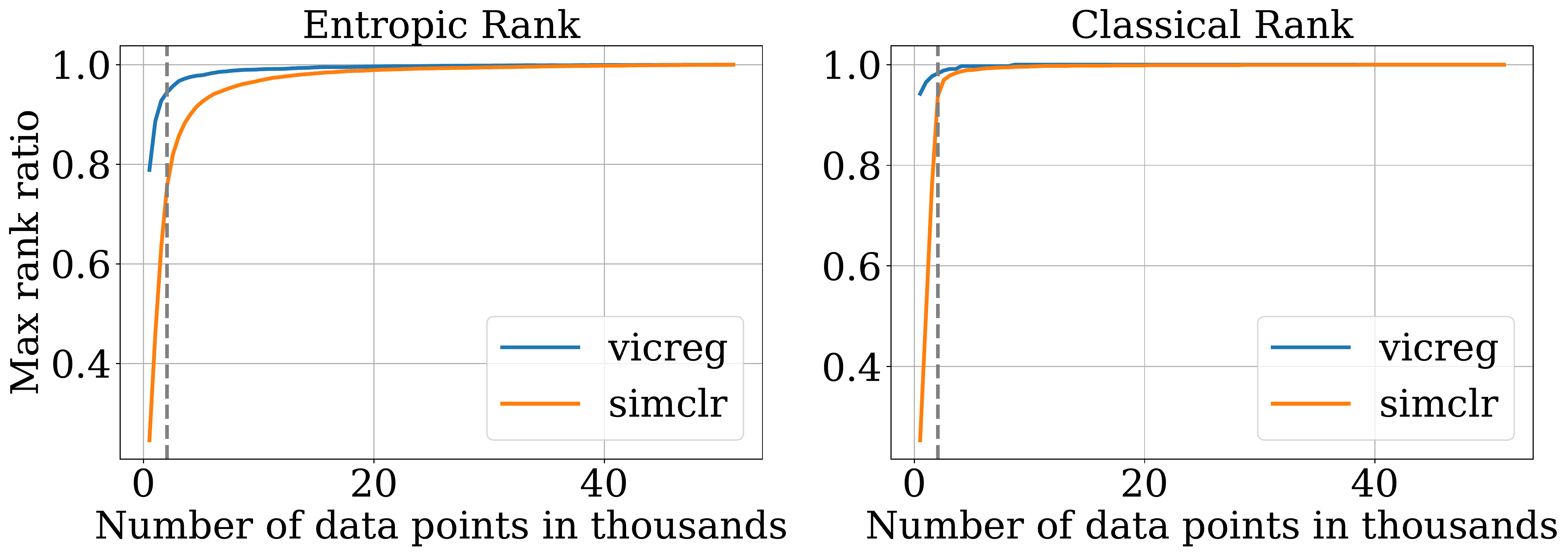}
    \caption{Convergence of the rank estimators on ImageNet as a function of the number of samples for 2048 dimensional outputs, as indicated by the vertical line. }
    \label{fig:rank-convergence}
\end{figure}

As we can see in ~\cref{fig:rank-convergence}, the rank estimates converge extremely quickly, especially for VICReg. For both VICReg and SimCLR, $10 000$ samples are enough to obtain more than $95\%$ of the final rank. It is worth noting that the entropic rank estimator converges more slowly than the classical rank estimator, as it is more sensitive to the singular values. The fact that the rank can be approximated with few samples is encouraging for its use during training and not only as a measure of performance after pretraining.

\section{Reproduction of figures with the classical rank estimator}\label{sec:repro-rank}

\begin{figure}[t!]
    \centering
    \includegraphics[width=1\textwidth]{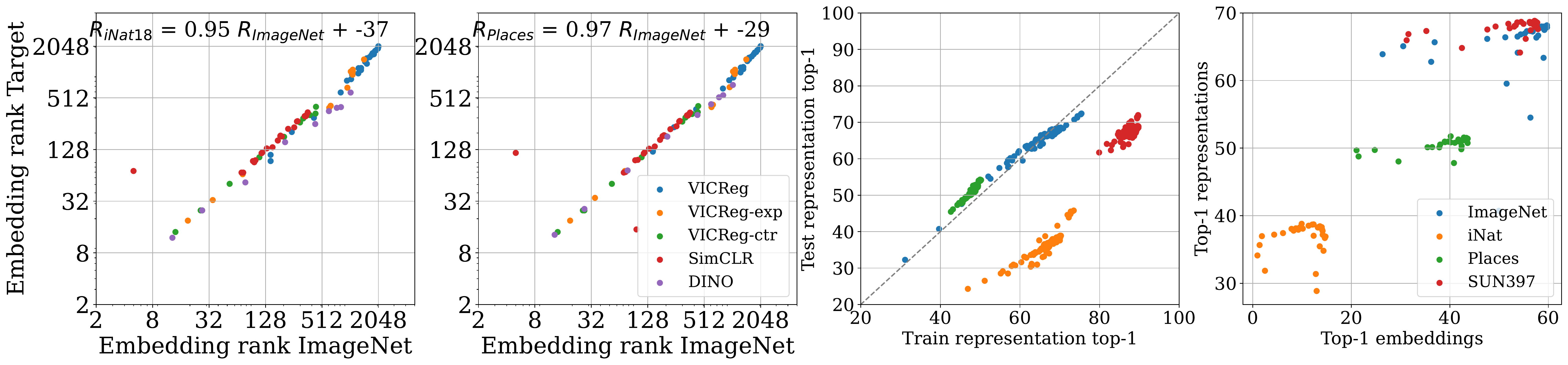}
    \caption{Reproduction of \cref{fig:transfer_collapse} with the classical rank estimator. Embeddings' rank transfers from source to target datasets. The estimates used $25 600$ images from the respective datasets.}
    \label{fig:transfer_collapse_2}
\end{figure}

\begin{figure}[t!]
    \centering
    \includegraphics[width=0.48\textwidth]{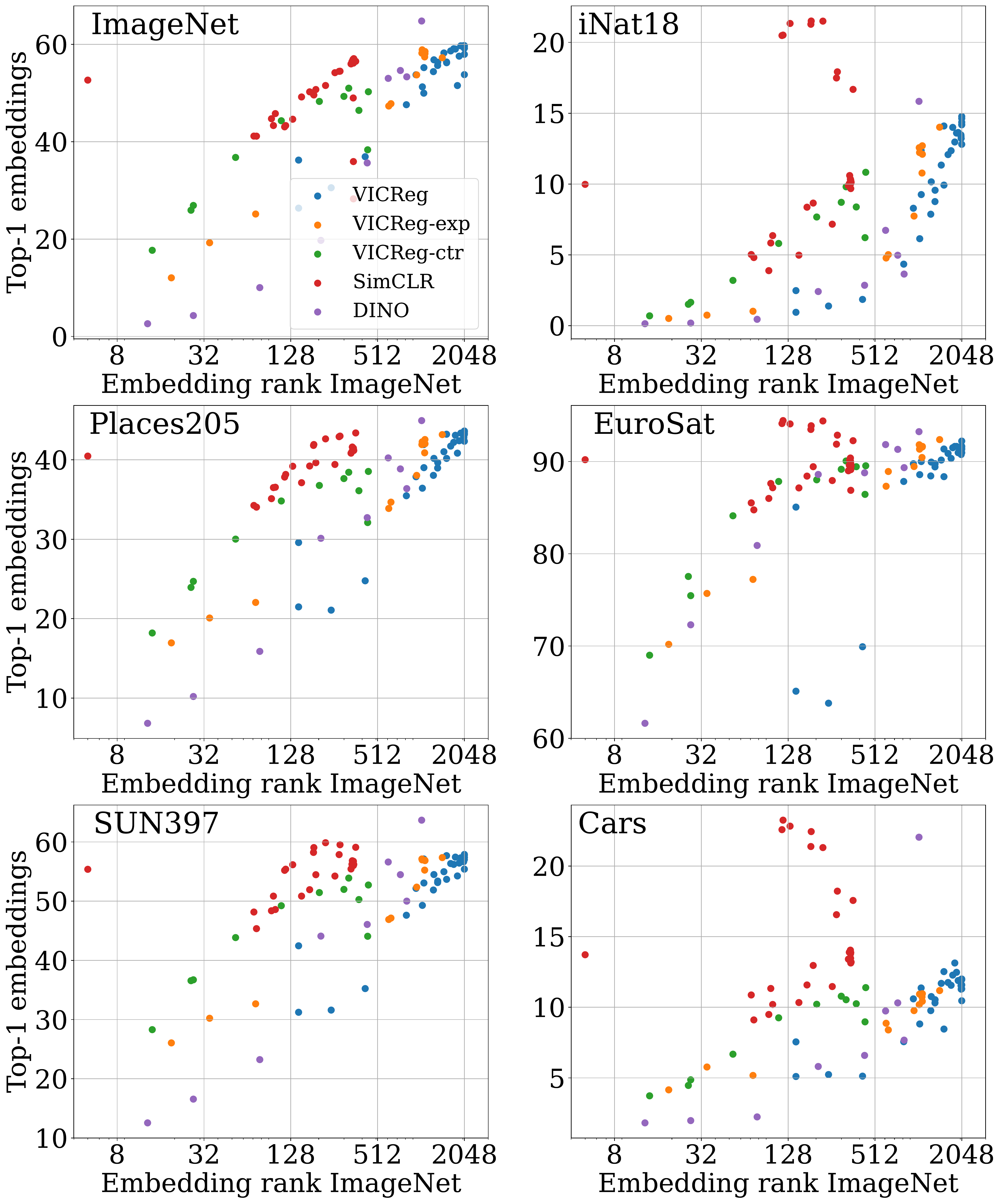}
    \hfill
    \includegraphics[width=0.48\textwidth]{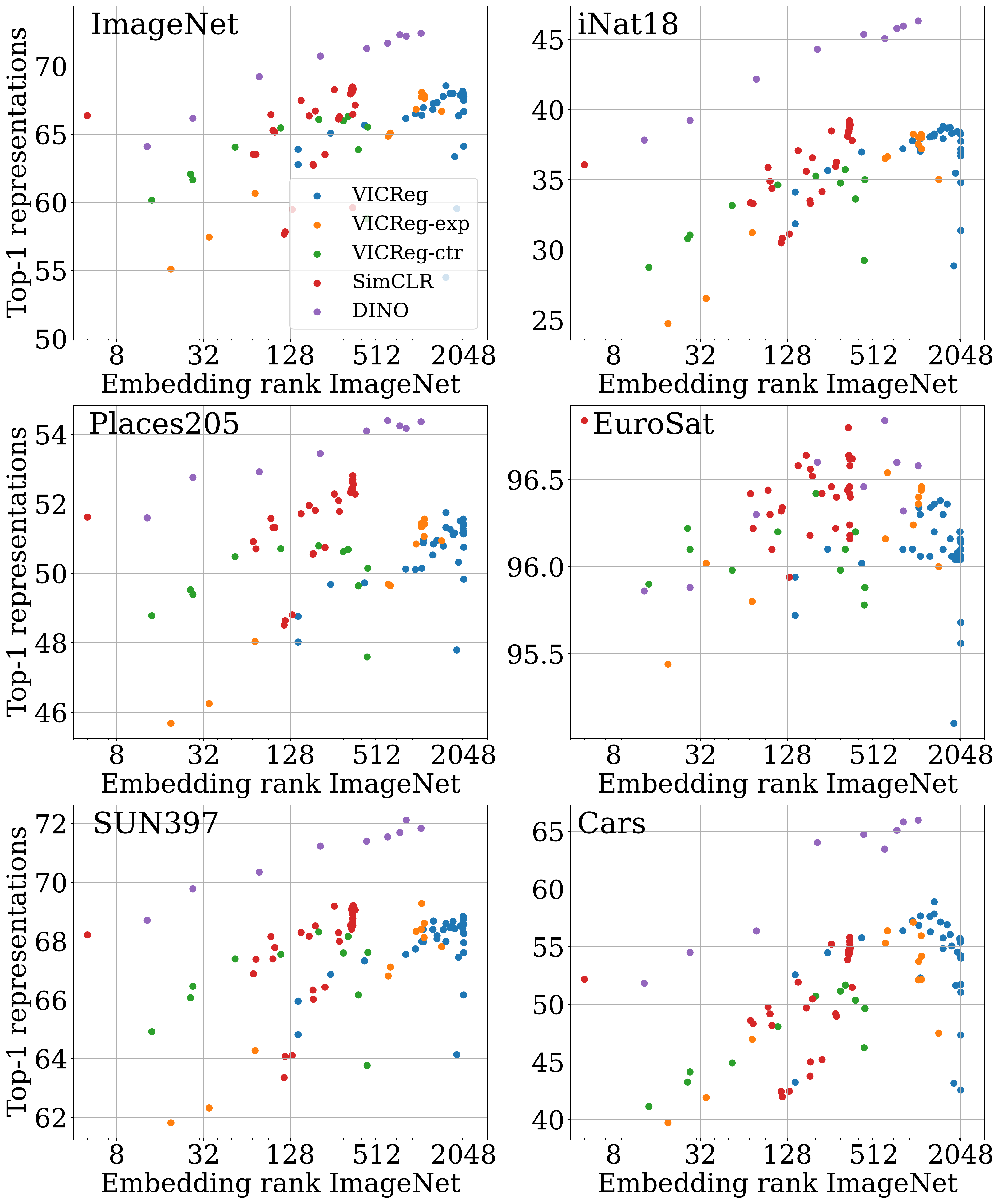}
    \caption{Reproduction of \cref{fig:perfs_repr} with the classical rank estimator.\textbf{(Left)} Validation of \name{} on embeddings, a higher ImageNet rank leads to improved performance across methods and datasets.\textbf{(Right)} Validation of \name{} on representations, where the link is even clearer, reinforcing \name{}'s practical use.}
    \label{fig:perfs_repr_2}
\end{figure}

As can be seen in \cref{fig:perfs_repr_2,fig:transfer_collapse_2}, the results that we obtain using the classical threshold-based rank estimator are extremely similar to the ones with the entropic estimator. The exact values do differ, but the behaviors stay the same. One of the main differences is illustrated in \cref{fig:perfs_repr_2}, where we can see that the target rank is almost identical to the source one when we previously saw a drop of around $50\%$. This can be explained by the fact that some features may be less present in the target dataset, reducing the associated singular values, and thus the entropic rank.\\
All of this shows that using one or the other will lead to similar results in practical scenarios.

\section{Detailed training and evaluation procedures}\label{sec:training-details}
\subsection{Pretraining}

\begin{table}[!h]
  \caption{Image augmentation parameters, taken from~\cite{grill2020byol}.}
  \label{tab:augmentation}
  \centering
  \begin{tabular}{lcc}
    \toprule
    Parameter & View 1 & View 2    \\
    \midrule
     Random crop probability             & $1.0$ & $1.0$ \\
     Horizontal flip probability         & $0.5$ & $0.5$ \\
     Color jittering probability         & $0.8$ & $0.8$ \\
     Brightness adjustment max intensity & $0.4$ & $0.4$ \\
     Contrast adjustment max intensity   & $0.4$ & $0.4$ \\
     Saturation adjustment max intensity & $0.2$ & $0.2$ \\
     Hue adjustment max intensity        & $0.1$ & $0.1$ \\
     Grayscale probability               & $0.2$ & $0.2$ \\
     Gaussian blurring probability       & $1.0$ & $0.1$ \\
     Solarization probability.           & $0.0$ & $0.2$ \\
    \bottomrule
  \end{tabular}
\end{table}

All pretrainings were done with ResNet-50 backbones. The projector used is a MLP with intermediate dimensions $8192,8192,2048$ ($8192,8192,2048,32768$ for DINO). VICReg, VICReg-ctr, VICReg-exp and SimCLR were trained with the LARS optimizer using a momentum of $0.9$, weight decay $10^{-6}$ and varying learning rates depending on the method. VICReg used $0.3$ base learning rate, SimCLR $0.5$ or $0.6$ depending on the experiment, VICReg-exp $0.6$ and VICReg-ctr $0.6$.
DINO was trained with AdamW~\citep{loshchilov2017adamw} using a learning rate of $0.00025$, using multi-crop 6 additional crops of size $96\times96$.
The learning rate is then computed as $lr=base\_lr*batch\_size/256$. We do a $10$-epochs linear warmup and then use cosine annealing. we use batch sizes of $2048$ for SimCLR and 1024 for other methods. SimCLR and VICReg-ctr also use a default temperature of $0.15$, and $0.1$ for VICReg-exp.\\
we use the image augmentation strategy from \cite{grill2020byol} illustrated in \cref{tab:augmentation}.
For the pretrainings on iNaturalist-18, we use the same protocol but with a 300 epoch pretraining to account for its smaller size compared to ImageNet.

\subsection{Evaluation}

\begin{table}[ht]
  \caption{Optimization parameters used to evaluate on downstream datasets}
  \label{tab:hp-eval}
  \centering

  \begin{tabular}{llcccc}
    \toprule
     Dataset & Optimizer & Weight decay & Momentum & Learning rate & Epochs\\
    \midrule
    ImageNet & SGD (w/ Nesterov) & $0.00004$ & $0.9$ & $0.3$ & $30$ \\
    iNaturalist18 & SGD (w/ Nesterov) & $0.0005$ & $0.9$ & $0.01$ & $84$ \\
    Places205 & SGD (w/ Nesterov) & $0.0005$ & $0.9$ & $0.01$ & $14$ \\
    EuroSat  & SGD (w/ Nesterov) & $0.0005$ & $0.9$ & $0.01$ & $28$\\
    Sun397   & SGD (w/ Nesterov) & $0.0005$ & $0.9$ & $0.01$ & $28$ \\
    StanfordCars & SGD (w/ Nesterov) & $0.0005$ & $0.9$ & $0.1$ & $28$ \\
    CIFAR10 & SGD (w/ Nesterov) & $0.0005$ & $0.9$ & $0.01$ & $28$ \\
    CIFAR100 & SGD (w/ Nesterov) & $0.0005$ & $0.9$ & $0.01$ & $28$ \\
    CLEVR-count & SGD (w/ Nesterov) & $0.0005$ & $0.9$ & $0.01$ & $50$ \\
    Food101 & SGD (w/ Nesterov) & $0.0005$ & $0.9$ & $0.01$ & $28$ \\
    VOC07 & \multicolumn{5}{c}{N/A, see in text}\\

    \bottomrule
  \end{tabular}
  
\end{table}

For all datasets except StanfordCars, we use the standard protocol in VISSL. On StanfordCars we mostly tuned the learning rate. The parameters that we use are described in \cref{tab:hp-eval}. For data augmentation, we use random resized crops and random horizontal flips during training, and center crop for evaluation. For VOC07, we follow the common protocol using SVMs, as used in~\cite{bardes2021vicreg}. We use the default VISSL settings for this evaluation.

\clearpage
\section{Detailed tables for hyperparameter selection}\label{sec:hp-tables}

\begin{table}[!h]
  \caption{Top-1 accuracies computed on representations when tuning hyperparameters with ImageNet validation performance, \name{} or with $\alpha$-ReQ.}
  \label{tab:hp-tuning-reprs-full}
  \centering
\begin{tabular}{llccccccccc}
    \toprule
     \multirow{2}{*}{Dataset} & \multirow{2}{*}{Method} &  \multicolumn{4}{c}{VICReg} &  \multicolumn{3}{c}{SimCLR} &  \multicolumn{2}{c}{DINO} \\ 
 \cmidrule(lr){3-6} \cmidrule(lr){7-9} \cmidrule(lr){10-11} &  & cov. & inv. & LR & WD & temp. & LR. & WD. & t-temp. & s-temp.\\ 
 \midrule 
\multirow{3}{*}{ImageNet} & ImageNet Oracle & $68.2$ & $68.2$ & $68.6$ & $68.0$ & $68.5$ & $68.5$ & $68.3$ & $72.3$ & $72.4$\\ 
 & \name{} & $67.8$ & $67.9$ & $68.2$ & $67.8$ & $67.1$ & $68.0$ & $68.3$ & $72.2$ & $72.4$\\ 
 & $\alpha$-ReQ & $67.9$ & $67.5$ & $59.5$ & $67.8$ & $63.5$ & $68.1$ & $32.3$ & $71.7$ & $66.2$\\ 
  \midrule 
\multirow{3}{*}{iNat18} & ImageNet Oracle & $38.4$ & $38.4$ & $38.8$ & $38.3$ & $39.2$ & $39.2$ & $38.9$ & $45.8$ & $46.3$\\ 
 & \name{} & $36.7$ & $37.2$ & $38.4$ & $38.3$ & $37.8$ & $38.1$ & $38.9$ & $46.0$ & $46.3$\\ 
 & $\alpha$-ReQ & $37.8$ & $36.9$ & $28.9$ & $38.3$ & $34.1$ & $38.4$ & $38.7$ & $45.1$ & $39.2$\\
\bottomrule
  \end{tabular}

\end{table}

\begin{table}[!h]
  \caption{Top-1 accuracies computed on embeddings when tuning hyperparameters with ImageNet validation performance, \name{} or with $\alpha$-ReQ.}
  \label{tab:hp-tuning-embs-full}
  \centering
 \begin{tabular}{llccccccccc}
    \toprule
     \multirow{2}{*}{Dataset} & \multirow{2}{*}{Method} &  \multicolumn{4}{c}{VICReg} &  \multicolumn{3}{c}{SimCLR} &  \multicolumn{2}{c}{DINO} \\ 
 \cmidrule(lr){3-6} \cmidrule(lr){7-9} \cmidrule(lr){10-11} &  & cov. & inv. & LR & WD & temp. & LR. & WD. & t-temp. & s-temp.\\ 
 \midrule 
\multirow{3}{*}{ImageNet} & ImageNet Oracle & $59.7$ & $59.7$ & $59.7$ & $59.7$ & $56.9$ & $56.9$ & $57.1$ & $54.6$ & $64.8$\\ 
 & \name{} & $59.6$ & $59.7$ & $59.7$ & $59.5$ & $56.5$ & $56.0$ & $57.1$ & $53.3$ & $64.8$\\ 
 & $\alpha$-ReQ & $59.6$ & $59.2$ & $36.2$ & $59.3$ & $51.5$ & $56.4$ & $49.0$ & $53.3$ & $53.3$\\ 
  \midrule 
\multirow{3}{*}{iNat18} & ImageNet Oracle & $13.5$ & $14.2$ & $13.5$ & $13.6$ & $10.3$ & $10.3$ & $10.1$ & $5.0$ & $15.8$\\ 
 & \name{} & $14.2$ & $14.2$ & $13.5$ & $13.4$ & $16.7$ & $9.9$ & $10.1$ & $3.6$ & $15.8$\\ 
 & $\alpha$-ReQ & $14.2$ & $14.8$ & $2.5$ & $13.2$ & $21.5$ & $10.0$ & $10.0$ & $3.6$ & $3.6$\\
\bottomrule
  \end{tabular}
 
\end{table}

\clearpage
\section{Complete performance tables}\label{sec:all-tables}

\begin{table}[!h]
  \caption{Hyperparameters for all runs.}
  \label{tab:all-hp-runs}
  \centering
  \begin{tabular}{llcccc}
    \toprule
     Method  &  Run & Batch size & Learning rate & Weight decay & Loss hyperparameters  \\
    \midrule
\multirow{32}{*}{VICReg} & 0 & $1024$ & $0.3$ & $10^{-6}$ &  $\lambda:25$, $\mu:25$, $\nu:0.3$ \\ 
 & 1 & $1024$ & $0.3$ & $10^{-6}$ &  $\lambda:25$, $\mu:25$, $\nu:0.4$ \\ 
 & 2 & $1024$ & $0.3$ & $10^{-6}$ &  $\lambda:25$, $\mu:25$, $\nu:0.5$ \\ 
 & 3 & $1024$ & $0.3$ & $10^{-6}$ &  $\lambda:25$, $\mu:25$, $\nu:0.6$ \\ 
 & 4 & $1024$ & $0.3$ & $10^{-6}$ &  $\lambda:25$, $\mu:25$, $\nu:0.7$ \\ 
 & 5 & $1024$ & $0.3$ & $10^{-6}$ &  $\lambda:25$, $\mu:25$, $\nu:0.8$ \\ 
 & 6 & $1024$ & $0.3$ & $10^{-6}$ &  $\lambda:25$, $\mu:25$, $\nu:0.9$ \\
\bottomrule
  \end{tabular}
\end{table}

\begin{table}[!h]
  \caption{Hyperparameters for all runs, continued.}
  \label{tab:all-hp-runs-2}
  \centering
  \begin{tabular}{llcccc}
    \toprule
     Method  &  Run & Batch size & Learning rate & Weight decay & Loss hyperparameters  \\
    \midrule
\multirow{11}{*}{VICReg-ctr} & 0 & $1024$ & $0.5$ & $10^{-6}$ &  $\lambda:1$, $\mu:1$, $\nu:1$, $\tau:0.05$ \\ 
 & 1 & $1024$ & $0.5$ & $10^{-6}$ &  $\lambda:1$, $\mu:1$, $\nu:1$, $\tau:0.07$ \\ 
 & 2 & $1024$ & $0.5$ & $10^{-6}$ &  $\lambda:1$, $\mu:1$, $\nu:1$, $\tau:0.1$ \\ 
 & 3 & $1024$ & $0.5$ & $10^{-6}$ &  $\lambda:1$, $\mu:1$, $\nu:1$, $\tau:0.2$ \\ 
 & 4 & $1024$ & $0.5$ & $10^{-6}$ &  $\lambda:1$, $\mu:1$, $\nu:1$, $\tau:0.3$ \\ 
 & 5 & $1024$ & $0.5$ & $10^{-6}$ &  $\lambda:1$, $\mu:1$, $\nu:1$, $\tau:0.4$ \\ 
 & 6 & $1024$ & $0.5$ & $10^{-6}$ &  $\lambda:1$, $\mu:1$, $\nu:0.1$, $\tau:0.1$ \\
\bottomrule
  \end{tabular}
\end{table}

\begin{table}[!h]
  \caption{Hyperparameters for all runs, continued.}
  \label{tab:all-hp-runs-3}
  \centering
  \begin{tabular}{llcccc}
    \toprule
     Method  &  Run & Batch size & Learning rate & Weight decay & Loss hyperparameters  \\
    \midrule
\multirow{9}{*}{DINO} & 0 & $1024$ & $2.5\times 10^{-4}$ & $10^{-6}$ &  $\tau_t:0.01$, \\ 
 & 1 & $1024$ & $2.5\times 10^{-4}$ & $10^{-6}$ &  $\tau_t:0.02$, \\ 
 & 2 & $1024$ & $2.5\times 10^{-4}$ & $10^{-6}$ &  $\tau_t:0.04$, \\ 
 & 3 & $1024$ & $2.5\times 10^{-4}$ & $10^{-6}$ &  $\tau_t:0.06$, \\ 
 & 4 & $1024$ & $2.5\times 10^{-4}$ & $10^{-6}$ &  $\tau_t:0.07$, \\ 
 & 5 & $1024$ & $2.5\times 10^{-4}$ & $10^{-6}$ &  $\tau_t:0.04$, $\tau_s:0.07$, \\ 
 & 6 & $1024$ & $2.5\times 10^{-4}$ & $10^{-6}$ &  $\tau_t:0.04$, $\tau_s:0.2$, \\ 
 & 7 & $1024$ & $2.5\times 10^{-4}$ & $10^{-6}$ &  $\tau_t:0.04$, $\tau_s:0.3$, \\ 
 & 8 & $1024$ & $2.5\times 10^{-4}$ & $10^{-6}$ &  $\tau_t:0.04$, $\tau_s:0.4$, \\ 
    \bottomrule
  \end{tabular}
\end{table}

\begin{table}[!h]
  \caption{Top-1 on representations in all settings.}
  \label{tab:all-perfs-reprs}
  \centering
  \begin{tabular}{llccccccc}
    \toprule
     Method  &  Run & ImageNet & iNat18 & Places205 & EuroSat & SUN397 & Cars \\
    \midrule
\multirow{32}{*}{VICReg} &0 & $63.90$ & $34.12$ & $48.77$ & $95.94$ & $65.96$ & $52.56$\\ 
 &1 & $65.08$ & $35.65$ & $49.68$ & $96.10$ & $66.87$ & $54.47$\\ 
 &2 & $65.67$ & $36.97$ & $49.73$ & $96.02$ & $67.33$ & $55.76$\\ 
 &3 & $66.17$ & $37.20$ & $50.13$ & $96.10$ & $67.55$ & $56.37$\\ 
 &4 & $66.40$ & $37.42$ & $50.15$ & $96.34$ & $67.99$ & $56.86$\\ 
 &5 & $66.83$ & $38.05$ & $50.53$ & $96.06$ & $68.40$ & $57.63$\\ 
 &6 & $67.30$ & $38.13$ & $50.96$ & $96.20$ & $68.08$ & $57.83$\\
\bottomrule
  \end{tabular}
\end{table}

\begin{table}[!h]
  \caption{Top-1 on representations in all settings, continued.}
  \label{tab:all-perfs-reprs-2}
  \centering
  \begin{tabular}{llccccccc}
    \toprule
     Method  &  Run & ImageNet & iNat18 & Places205 & EuroSat & SUN397 & Cars \\
    \midrule
\multirow{11}{*}{VICReg-ctr} &0 & $65.54$ & $35.00$ & $50.15$ & $95.88$ & $67.62$ & $49.63$\\ 
 &1 & $66.32$ & $35.72$ & $50.69$ & $96.10$ & $68.16$ & $51.66$\\ 
 &2 & $66.09$ & $35.26$ & $50.80$ & $96.42$ & $68.32$ & $50.72$\\ 
 &3 & $64.06$ & $33.16$ & $50.48$ & $95.98$ & $67.40$ & $44.91$\\ 
 &4 & $62.06$ & $30.80$ & $49.53$ & $96.22$ & $66.08$ & $43.24$\\ 
 &5 & $60.17$ & $28.76$ & $48.78$ & $95.90$ & $64.92$ & $41.13$\\ 
 &6 & $61.66$ & $31.05$ & $49.40$ & $96.10$ & $66.47$ & $44.12$\\
\bottomrule
  \end{tabular}
\end{table}

\begin{table}[!h]
  \caption{Top-1 on representations in all settings, continued.}
  \label{tab:all-perfs-reprs-3}
  \centering
  \begin{tabular}{llccccccc}
    \toprule
     Method  &  Run & ImageNet & iNat18 & Places205 & EuroSat & SUN397 & Cars \\
    \midrule
\multirow{9}{*}{DINO} &0 & $70.74$ & $44.30$ & $53.45$ & $96.60$ & $71.23$ & $64.03$\\ 
 &1 & $71.29$ & $45.37$ & $54.10$ & $96.46$ & $71.40$ & $64.73$\\ 
 &2 & $72.19$ & $45.96$ & $54.18$ & $96.32$ & $72.12$ & $65.81$\\ 
 &3 & $72.30$ & $45.80$ & $54.25$ & $96.60$ & $71.69$ & $65.09$\\ 
 &4 & $71.68$ & $45.06$ & $54.40$ & $96.84$ & $71.55$ & $63.46$\\ 
 &5 & $72.41$ & $46.32$ & $54.37$ & $96.58$ & $71.84$ & $65.97$\\ 
 &6 & $69.23$ & $42.18$ & $52.93$ & $96.30$ & $70.35$ & $56.36$\\ 
 &7 & $66.18$ & $39.24$ & $52.77$ & $95.88$ & $69.78$ & $54.48$\\ 
 &8 & $64.10$ & $37.83$ & $51.60$ & $95.86$ & $68.71$ & $51.82$\\ 
    \bottomrule
  \end{tabular}
\end{table}

\begin{table}[!h]
  \caption{Top-1 on representations in all settings, continued.}
  \label{tab:all-perfs-reprs-2-1}
  \centering
  \begin{tabular}{llccccccc}
    \toprule
     Method  &  Run & ImageNet & CIFAR10 & CIFAR100 & FOOD101 & VOC07 & CLEVR-count \\
    \midrule
\multirow{32}{*}{VICReg} &0 & $63.90$ & $88.94$ & $69.92$ & $75.00$ & $81.49$ & $48.94$\\ 
 &1 & $65.08$ & $88.74$ & $69.96$ & $75.00$ & $82.14$ & $49.88$\\ 
 &2 & $65.67$ & $88.25$ & $70.23$ & $75.00$ & $82.35$ & $54.96$\\ 
 &3 & $66.17$ & $89.17$ & $71.51$ & $75.00$ & $82.97$ & $52.35$\\ 
 &4 & $66.40$ & $89.41$ & $71.70$ & $75.01$ & $82.81$ & $55.27$\\ 
 &5 & $66.83$ & $89.91$ & $72.12$ & $75.00$ & $83.10$ & $55.95$\\ 
 &6 & $67.30$ & $90.11$ & $71.90$ & $75.01$ & $83.15$ & $54.37$\\
\bottomrule
  \end{tabular}
\end{table}

\begin{table}[!h]
  \caption{Top-1 on representations in all settings, continued.}
  \label{tab:all-perfs-reprs-2-2}
  \centering
  \begin{tabular}{llccccccc}
    \toprule
     Method  &  Run & ImageNet & CIFAR10 & CIFAR100 & FOOD101 & VOC07 & CLEVR-count \\
    \midrule
\multirow{10}{*}{VICReg-ctr} &0 & $65.54$ & $88.87$ & $70.77$ & $75.01$ & $83.28$ & $53.97$\\ 
 &1 & $66.32$ & $89.57$ & $70.93$ & $75.00$ & $84.17$ & $53.19$\\ 
 &2 & $66.09$ & $89.49$ & $71.17$ & $75.00$ & $83.90$ & $53.29$\\ 
 &3 & $64.06$ & $89.62$ & $71.39$ & $75.00$ & $83.18$ & $48.57$\\ 
 &4 & $62.06$ & $88.60$ & $69.41$ & $75.00$ & $82.35$ & $46.48$\\ 
 &5 & $60.17$ & $88.97$ & $68.61$ & $75.00$ & $81.43$ & $51.27$\\ 
 &6 & $65.47$ & $89.65$ & $71.62$ & $75.01$ & $84.09$ & $51.07$\\
\bottomrule
  \end{tabular}
\end{table}

\begin{table}[!h]
  \caption{Top-1 on representations in all settings, continued.}
  \label{tab:all-perfs-reprs-2-3}
  \centering
  \begin{tabular}{llccccccc}
    \toprule
     Method  &  Run & ImageNet & CIFAR10 & CIFAR100 & FOOD101 & VOC07 & CLEVR-count \\
    \midrule
\multirow{9}{*}{DINO} &0 & $70.74$ & $91.87$ & $74.71$ & $75.00$ & $87.50$ & $51.67$\\ 
 &1 & $71.29$ & $91.81$ & $74.33$ & $75.00$ & $87.99$ & $57.19$\\ 
 &2 & $72.19$ & $90.51$ & $73.09$ & $75.01$ & $88.35$ & $56.78$\\ 
 &3 & $72.30$ & $91.46$ & $74.34$ & $75.00$ & $88.39$ & $56.75$\\ 
 &4 & $71.68$ & $90.89$ & $72.82$ & $75.00$ & $88.48$ & $59.19$\\ 
 &5 & $72.41$ & $92.24$ & $75.58$ & $75.00$ & $88.04$ & $59.29$\\ 
 &6 & $69.23$ & $90.36$ & $71.86$ & $75.00$ & $87.61$ & $53.87$\\ 
 &7 & $66.18$ & $87.97$ & $68.18$ & $75.00$ & $86.72$ & $54.56$\\ 
 &8 & $64.10$ & $86.80$ & $66.68$ & $75.01$ & $85.38$ & $55.23$\\ 
    \bottomrule
  \end{tabular}
\end{table}

\begin{table}[!h]
  \caption{Top-1 on embeddings in all settings.}
  \label{tab:all-perfs-embs}
  \centering
  \begin{tabular}{llccccccc}
    \toprule
     Method  &  Run & ImageNet & iNat18 & Places205 & EuroSat & SUN397 & Cars \\
    \midrule
\multirow{32}{*}{VICReg} &0 & $26.35$ & $0.95$ & $21.48$ & $65.10$ & $31.23$ & $5.10$\\ 
 &1 & $30.54$ & $1.39$ & $21.07$ & $63.80$ & $31.60$ & $5.24$\\ 
 &2 & $36.92$ & $1.85$ & $24.77$ & $69.92$ & $35.24$ & $5.12$\\ 
 &3 & $47.60$ & $4.34$ & $35.48$ & $87.84$ & $47.62$ & $7.56$\\ 
 &4 & $51.26$ & $6.14$ & $36.43$ & $88.58$ & $49.29$ & $8.82$\\ 
 &5 & $54.39$ & $7.87$ & $38.04$ & $88.44$ & $51.88$ & $9.76$\\ 
 &6 & $55.66$ & $8.76$ & $38.97$ & $89.42$ & $53.14$ & $10.31$\\
\bottomrule
  \end{tabular}
\end{table}

\begin{table}[!h]
  \caption{Top-1 on embeddings in all settings, continued.}
  \label{tab:all-perfs-embs-2}
  \centering
  \begin{tabular}{llccccccc}
    \toprule
     Method  &  Run & ImageNet & iNat18 & Places205 & EuroSat & SUN397 & Cars \\
    \midrule
\multirow{11}{*}{VICReg-ctr} &0 & $50.26$ & $10.83$ & $38.54$ & $89.54$ & $52.73$ & $11.39$\\ 
 &1 & $50.99$ & $9.81$ & $38.43$ & $90.06$ & $53.90$ & $10.53$\\ 
 &2 & $48.27$ & $7.67$ & $36.78$ & $88.02$ & $51.45$ & $10.21$\\ 
 &3 & $36.77$ & $3.20$ & $30.01$ & $84.12$ & $43.85$ & $6.68$\\ 
 &4 & $25.92$ & $1.51$ & $23.93$ & $77.54$ & $36.57$ & $4.46$\\ 
 &5 & $17.69$ & $0.70$ & $18.19$ & $69.00$ & $28.30$ & $3.73$\\ 
 &6 & $26.90$ & $1.65$ & $24.69$ & $75.46$ & $36.72$ & $4.86$\\
\bottomrule
  \end{tabular}
\end{table}

\begin{table}[!h]
  \caption{Top-1 on embeddings in all settings, continued.}
  \label{tab:all-perfs-embs-3}
  \centering
  \begin{tabular}{llccccccc}
    \toprule
     Method  &  Run & ImageNet & iNat18 & Places205 & EuroSat & SUN397 & Cars \\
    \midrule
\multirow{9}{*}{DINO} &0 & $19.72$ & $2.41$ & $30.12$ & $88.60$ & $44.10$ & $5.81$\\ 
 &1 & $35.64$ & $2.85$ & $32.72$ & $88.78$ & $46.07$ & $6.59$\\ 
 &2 & $53.33$ & $3.64$ & $36.38$ & $89.34$ & $50.00$ & $7.67$\\ 
 &3 & $54.63$ & $4.97$ & $38.86$ & $91.32$ & $54.48$ & $10.31$\\ 
 &4 & $53.01$ & $6.73$ & $40.24$ & $91.84$ & $56.61$ & $9.74$\\ 
 &5 & $64.79$ & $15.84$ & $44.95$ & $93.24$ & $63.68$ & $22.04$\\ 
 &6 & $10.02$ & $0.45$ & $15.87$ & $80.90$ & $23.24$ & $2.24$\\ 
 &7 & $4.28$ & $0.18$ & $10.19$ & $72.30$ & $16.56$ & $1.98$\\ 
 &8 & $2.60$ & $0.14$ & $6.81$ & $61.62$ & $12.55$ & $1.82$\\  

    \bottomrule
  \end{tabular}
\end{table}

\begin{table}[!h]
  \caption{Top-1 on embeddings in all settings, continued.}
  \label{tab:all-perfs-embs-2-1}
  \centering
  \begin{tabular}{llccccccc}
    \toprule
     Method  &  Run & ImageNet & CIFAR10 & CIFAR100 & FOOD101 & VOC07 & CLEVR-count \\
    \midrule
\multirow{32}{*}{VICReg} &0 & $26.35$ & $59.23$ & $25.84$ & $75.00$ & $66.15$ & $21.75$\\ 
 &1 & $30.54$ & $60.57$ & $25.67$ & $75.01$ & $67.33$ & $19.64$\\ 
 &2 & $36.92$ & $63.77$ & $29.67$ & $75.00$ & $70.94$ & $23.72$\\ 
 &3 & $47.60$ & $75.59$ & $44.58$ & $75.01$ & $75.90$ & $41.44$\\ 
 &4 & $51.26$ & $76.88$ & $45.26$ & $75.01$ & $76.85$ & $34.20$\\ 
 &5 & $54.39$ & $78.34$ & $49.63$ & $75.01$ & $77.85$ & $40.24$\\ 
 &6 & $55.66$ & $78.71$ & $49.89$ & $75.01$ & $78.17$ & $38.37$\\
\bottomrule
  \end{tabular}
\end{table}

\begin{table}[!h]
  \caption{Top-1 on embeddings in all settings, continued.}
  \label{tab:all-perfs-embs-2-2}
  \centering
  \begin{tabular}{llccccccc}
    \toprule
     Method  &  Run & ImageNet & CIFAR10 & CIFAR100 & FOOD101 & VOC07 & CLEVR-count \\
    \midrule
\multirow{10}{*}{VICReg-ctr} &0 & $50.26$ & $78.76$ & $49.34$ & $75.00$ & $77.89$ & $37.61$\\ 
 &1 & $50.99$ & $78.63$ & $49.80$ & $75.00$ & $78.77$ & $43.76$\\ 
 &2 & $48.27$ & $77.86$ & $48.75$ & $75.00$ & $78.48$ & $40.49$\\ 
 &3 & $36.77$ & $73.83$ & $40.44$ & $75.00$ & $72.19$ & $38.16$\\ 
 &4 & $25.92$ & $66.81$ & $32.82$ & $75.00$ & $65.32$ & $31.63$\\ 
 &5 & $17.69$ & $63.94$ & $25.50$ & $75.00$ & $58.48$ & $31.40$\\ 
 &6 & $44.31$ & $76.82$ & $46.43$ & $75.00$ & $77.16$ & $42.48$\\
\bottomrule
  \end{tabular}
\end{table}
\begin{table}[!h]
  \caption{Top-1 on embeddings in all settings, continued.}
  \label{tab:all-perfs-embs-2-3}
  \centering
  \begin{tabular}{llccccccc}
    \toprule
     Method  &  Run & ImageNet & CIFAR10 & CIFAR100 & FOOD101 & VOC07 & CLEVR-count \\
    \midrule
\multirow{9}{*}{DINO} &0 & $19.72$ & $81.24$ & $53.97$ & $75.00$ & $77.21$ & $42.39$\\ 
 &1 & $35.64$ & $81.66$ & $55.32$ & $75.00$ & $82.48$ & $45.23$\\ 
 &2 & $53.33$ & $80.70$ & $53.03$ & $75.00$ & $84.22$ & $41.93$\\ 
 &3 & $54.63$ & $84.26$ & $58.49$ & $75.00$ & $85.25$ & $49.89$\\ 
 &4 & $53.01$ & $83.20$ & $58.14$ & $75.00$ & $85.71$ & $48.72$\\ 
 &5 & $64.79$ & $87.02$ & $65.43$ & $75.00$ & $87.03$ & $51.19$\\ 
 &6 & $10.02$ & $70.98$ & $37.66$ & $75.00$ & $61.27$ & $32.83$\\ 
 &7 & $4.28$ & $59.02$ & $22.71$ & $75.00$ & $50.02$ & $34.77$\\ 
 &8 & $2.60$ & $47.97$ & $16.26$ & $75.00$ & $41.16$ & $27.13$\\  

    \bottomrule
  \end{tabular}
\end{table}

\begin{table}[!h]
  \caption{Rank after projector in all settings.}
  \label{tab:all-ranks}
  \centering
  \begin{tabular}{llccccccc}
    \toprule
     Method  &  Run & ImageNet & iNat18 & Places205 & EuroSat & SUN397 & Cars \\
    \midrule
\multirow{32}{*}{VICReg} &0 & $102.07$ & $38.10$ & $44.39$ & $14.61$ & $32.40$ & $7.03$\\ 
 &1 & $229.81$ & $92.53$ & $129.47$ & $88.78$ & $98.44$ & $12.58$\\ 
 &2 & $374.25$ & $135.79$ & $206.29$ & $120.31$ & $163.31$ & $19.77$\\ 
 &3 & $612.12$ & $261.34$ & $336.16$ & $228.60$ & $265.64$ & $38.90$\\ 
 &4 & $831.49$ & $382.55$ & $467.68$ & $366.78$ & $374.50$ & $59.15$\\ 
 &5 & $952.55$ & $449.44$ & $539.24$ & $428.87$ & $435.94$ & $77.36$\\ 
 &6 & $1033.93$ & $493.50$ & $587.19$ & $477.69$ & $478.34$ & $88.28$\\
\bottomrule
  \end{tabular}
\end{table}

\begin{table}[!h]
  \caption{Rank after projector in all settings, continued.}
  \label{tab:all-ranks-3}
  \centering
  \begin{tabular}{llccccccc}
    \toprule
     Method  &  Run & ImageNet & iNat18 & Places205 & EuroSat & SUN397 & Cars \\
    \midrule
\multirow{11}{*}{VICReg-ctr} &0 & $382.33$ & $224.68$ & $252.33$ & $207.81$ & $220.68$ & $69.48$\\ 
 &1 & $278.88$ & $163.91$ & $183.32$ & $154.70$ & $160.71$ & $50.29$\\ 
 &2 & $169.33$ & $101.44$ & $114.49$ & $97.89$ & $99.84$ & $34.97$\\ 
 &3 & $48.47$ & $32.38$ & $34.93$ & $32.77$ & $31.76$ & $12.53$\\ 
 &4 & $23.22$ & $16.72$ & $17.90$ & $17.70$ & $16.63$ & $7.38$\\ 
 &5 & $12.88$ & $10.03$ & $10.31$ & $10.66$ & $9.71$ & $5.01$\\ 
 &6 & $22.96$ & $16.87$ & $17.77$ & $17.30$ & $16.55$ & $7.61$\\
\bottomrule
  \end{tabular}
\end{table}

\begin{table}[!h]
  \caption{Rank after projector in all settings, continued.}
  \label{tab:all-ranks-2}
  \centering
  \begin{tabular}{llccccccc}
    \toprule
     Method  &  Run & ImageNet & iNat18 & Places205 & EuroSat & SUN397 & Cars \\
    \midrule
\multirow{9}{*}{DINO} &0 & $150.51$ & $62.45$ & $106.14$ & $100.08$ & $90.46$ & $23.52$\\ 
 &1 & $276.35$ & $110.16$ & $179.46$ & $163.25$ & $147.50$ & $36.81$\\ 
 &2 & $482.51$ & $190.78$ & $291.91$ & $213.05$ & $236.76$ & $52.59$\\ 
 &3 & $409.79$ & $180.78$ & $269.96$ & $196.90$ & $225.47$ & $70.37$\\ 
 &4 & $347.69$ & $168.61$ & $235.11$ & $168.90$ & $202.51$ & $55.07$\\ 
 &5 & $523.67$ & $330.31$ & $392.16$ & $271.77$ & $356.99$ & $155.13$\\ 
 &6 & $61.59$ & $21.32$ & $39.89$ & $34.88$ & $32.55$ & $2.52$\\ 
 &7 & $22.43$ & $8.26$ & $17.91$ & $9.86$ & $15.62$ & $1.93$\\ 
 &8 & $11.01$ & $5.84$ & $8.27$ & $8.08$ & $7.83$ & $1.39$\\ 
    \bottomrule
  \end{tabular}
\end{table}

\begin{table}[!h]
  \caption{Rank after projector in all settings, continued.}
  \label{tab:all-ranks-2-1}
  \centering
  \begin{tabular}{llccccccc}
    \toprule
     Method  &  Run & ImageNet & CIFAR10 & CIFAR100 & FOOD101 & VOC07 & CLEVR-count \\
    \midrule
\multirow{32}{*}{VICReg} &0 & $102.07$ & $38.10$ & $44.39$ & $14.61$ & $32.40$ & $7.03$\\ 
 &1 & $229.81$ & $92.53$ & $129.47$ & $88.78$ & $98.44$ & $12.58$\\ 
 &2 & $374.25$ & $135.79$ & $206.29$ & $120.31$ & $163.31$ & $19.77$\\ 
 &3 & $612.12$ & $261.34$ & $336.16$ & $228.60$ & $265.64$ & $38.90$\\ 
 &4 & $831.49$ & $382.55$ & $467.68$ & $366.78$ & $374.50$ & $59.15$\\ 
 &5 & $952.55$ & $449.44$ & $539.24$ & $428.87$ & $435.94$ & $77.36$\\ 
 &6 & $1033.93$ & $493.50$ & $587.19$ & $477.69$ & $478.34$ & $88.28$\\
\bottomrule
  \end{tabular}
\end{table}

\begin{table}[!h]
  \caption{Rank after projector in all settings, continued.}
  \label{tab:all-ranks-2-2}
  \centering
  \begin{tabular}{llccccccc}
    \toprule
     Method  &  Run & ImageNet & CIFAR10 & CIFAR100 & FOOD101 & VOC07 & CLEVR-count \\
    \midrule
\multirow{10}{*}{VICReg-ctr} &0 & $382.33$ & $224.68$ & $252.33$ & $207.81$ & $220.68$ & $69.48$\\ 
 &1 & $278.88$ & $163.91$ & $183.32$ & $154.70$ & $160.71$ & $50.29$\\ 
 &2 & $169.33$ & $101.44$ & $114.49$ & $97.89$ & $99.84$ & $34.97$\\ 
 &3 & $48.47$ & $32.38$ & $34.93$ & $32.77$ & $31.76$ & $12.53$\\ 
 &4 & $23.22$ & $16.72$ & $17.90$ & $17.70$ & $16.63$ & $7.38$\\ 
 &5 & $12.88$ & $10.03$ & $10.31$ & $10.66$ & $9.71$ & $5.01$\\ 
 &6 & $96.33$ & $62.08$ & $68.05$ & $60.68$ & $60.39$ & $22.59$\\
\bottomrule
  \end{tabular}
\end{table}

\begin{table}[!h]
  \caption{Rank after projector in all settings, continued.}
  \label{tab:all-ranks-2-3}
  \centering
  \begin{tabular}{llccccccc}
    \toprule
     Method  &  Run & ImageNet & CIFAR10 & CIFAR100 & FOOD101 & VOC07 & CLEVR-count \\
    \midrule
\multirow{9}{*}{DINO} &0 & $150.51$ & $62.45$ & $106.14$ & $100.08$ & $90.46$ & $23.52$\\ 
 &1 & $276.35$ & $110.16$ & $179.46$ & $163.25$ & $147.50$ & $36.81$\\ 
 &2 & $482.51$ & $190.78$ & $291.91$ & $213.05$ & $236.76$ & $52.59$\\ 
 &3 & $409.79$ & $180.78$ & $269.96$ & $196.90$ & $225.47$ & $70.37$\\ 
 &4 & $347.69$ & $168.61$ & $235.11$ & $168.90$ & $202.51$ & $55.07$\\ 
 &5 & $523.67$ & $330.31$ & $392.16$ & $271.77$ & $356.99$ & $155.13$\\ 
 &6 & $61.59$ & $21.32$ & $39.89$ & $34.88$ & $32.55$ & $2.52$\\ 
 &7 & $22.43$ & $8.26$ & $17.91$ & $9.86$ & $15.62$ & $1.93$\\ 
 &8 & $11.01$ & $5.84$ & $8.27$ & $8.08$ & $7.83$ & $1.39$\\ 
    \bottomrule
  \end{tabular}
\end{table}

\end{document}